\documentclass{article}
\usepackage{import}
\usepackage{microtype}
\usepackage{graphicx}
\usepackage{subfigure}
\usepackage{booktabs} % for professional tables

\usepackage{hyperref}

\usepackage[accepted]{icml2023}

\usepackage{amsmath}
\usepackage{amssymb}
\usepackage{mathtools}
\usepackage{amsthm}

\usepackage[capitalize,noabbrev]{cleveref}

\theoremstyle{plain}
\newtheorem{theorem}{Theorem}[section]

\theoremstyle{definition}
\newtheorem{definition}[theorem]{Definition}

\theoremstyle{remark}

\usepackage{overpic}
\usepackage{siunitx}
\newcommand{\expnumber}[2]{{#1}\mathrm{e}{#2}}

\usepackage[textsize=tiny]{todonotes}

\icmltitlerunning{Data Representations' Study of Latent Image Manifolds}

\begin{document}

\twocolumn[
\icmltitle{Data Representations' Study of Latent Image Manifolds}



\icmlsetsymbol{equal}{*}

\begin{icmlauthorlist}
\icmlauthor{Ilya Kaufman}{sch}
\icmlauthor{Omri Azencot}{sch}
\end{icmlauthorlist}

\icmlaffiliation{sch}{Department of Computer Science, Ben-Gurion University of the Negev, Beer-Sheva, Israel}

\icmlcorrespondingauthor{Ilya Kaufman}{ilyakau@post.bgu.ac.il}
% \icmlcorrespondingauthor{Omri Azencot}{azencot@cs.bgu.ac.il}

\icmlkeywords{Machine Learning, ICML}

\vskip 0.3in
]



\printAffiliationsAndNotice{}  % leave blank if no need to mention equal contribution

\begin{abstract}
Deep neural networks have been demonstrated to achieve phenomenal success in many domains, and yet their inner mechanisms are not well understood. In this paper, we investigate the \emph{curvature} of image manifolds, i.e., the manifold deviation from being flat in its principal directions. We find that state-of-the-art trained convolutional neural networks for image classification have a characteristic curvature profile along layers: an initial steep increase, followed by a long phase of a plateau, and followed by another increase. In contrast, this behavior does not appear in untrained networks in which the curvature flattens. We also show that the curvature gap between the last two layers has a strong correlation with the generalization capability of the network. Moreover, we find that the intrinsic dimension of latent codes is not necessarily indicative of curvature. Finally, we observe that common regularization methods such as mixup yield flatter representations when compared to other methods. Our experiments show consistent results over a variety of deep learning architectures and multiple data sets. Our code is publicly available at \url{https://github.com/azencot-group/CRLM}
\end{abstract}

\vspace{-4mm}

\section{Introduction}
\label{Sec:Intro}

% complex data, manifold hypothesis, DL works great, no understanding; TODO: add missing references, replace ref to ml features??
Real-world data arising from scientific and engineering problems is often high-dimensional and complex. Using such data for downstream tasks may seem hopeless at first glance. Nevertheless, the widely accepted \emph{manifold hypothesis}~\cite{cayton2005algorithms} stating that complex high-dimensional data is intrinsically low-dimensional, suggests that not all hope is lost. Indeed, significant efforts in machine learning~\cite{khalid2014survey, bengio2013representation} have been dedicated to developing tools for extracting meaningful low-dimensional features from real-world information. Particularly successful in several challenging tasks such as classification~\cite{krizhevsky2017imagenet} and recognition~\cite{girshick2014rich} are deep learning approaches which manipulate data via nonlinear neural networks. Unfortunately, the inner mechanisms of deep models are not well understood at large.

% manifold learning, intrinsic dimension, hunchback profile, ID vs ACC
Motivated by the manifold hypothesis and more generally, manifold learning~\cite{belkin2003laplacian}, several recent approaches proposed to analyze deep models by their latent representations. A manifold is a topological space locally similar to a Euclidean domain at each of its points~\cite{lee2013smooth}. A key property of a manifold is its \emph{intrinsic dimension}, defined as the dimension of the related Euclidean domain. Recent studies estimated the intrinsic dimension (ID) along layers of trained neural networks using neighborhood information~\cite{ansuini:NIPS:2019:intrinsic} and topological data analysis~\cite{birdal2021intrinsic}. Remarkably, it has been shown that the ID admits a characteristic ``hunchback'' profile~\cite{ansuini:NIPS:2019:intrinsic}, i.e., it increases in the first layers and then it decreases progressively. Moreover, the ID was found to be strongly correlated with the network performance.

% ID single number, Riemann manifold, examples: distances & curvature, conjecture of curvature, what is curvature, 
Still, the intrinsic dimension is only a single measure, providing limited knowledge of the manifold. To consider other properties, the manifold has to be equipped with an additional structure. In this work, we focus on Riemannian manifolds which are differentiable manifolds with an inner product \cite{lee2006riemannian}. Riemannian manifolds can be described using properties such as angles, distances, and curvatures. For instance, the curvature in two dimensions is the amount by which a surface deviates from being a plane, which is completely flat. \citet{ansuini:NIPS:2019:intrinsic} conjectured that while the intrinsic dimension decreases with network depth, the underlying manifold is highly curved. Our study confirms the latter conjecture empirically by estimating the \emph{principal curvatures} of latent representations of popular deep convolutional classification models trained on benchmark datasets.

% existing curvature work, systematic, our approach, several arch. and datasets; TODO: want to mention complexity/efficiency of curvature computations?
Previously, curvature estimates were used in the analysis of trained deep models to compare between two neural networks~\cite{yu2018curvature}, and to explore the decision boundary profile of classification models~\cite{kaul2019riemannian}. However, there has not been an extensive and systematic investigation that characterizes the curvature profile of data representations along layers of deep neural networks, similar to existing studies on the intrinsic dimension. In this paper, we take a step forward toward bridging this gap. To estimate principal curvatures per sample, we compute the eigenvalues of the manifold's Hessian, following the algorithm introduced in~\cite{li2018curvature}. Our evaluation focuses on convolutional neural network (CNN) architectures such as VGG~\cite{simonyan2014very} and ResNet~\cite{he2016deep} and on image classification benchmark datasets such as CIFAR-10 and CIFAR-100~\cite{krizhevsky2009learning}. We address the following questions:
\begin{itemize}
    \item How does curvature vary along the layers of CNNs? Do CNNs learn flat manifolds, or, alternatively, highly-curved data representations? How do common regularizers such as weight decay and mixup affect the curvature profile?
    \item Do curvature estimates of a trained network are indicative of its performance? Is there an indicator that generalize across different architectures and datasets?
    \item Is there a correlation between curvature and other geometric properties of the manifold, such as the intrinsic dimension? Can we deduce the curvature behavior along layers using dimensionality estimation tools? 
\end{itemize}

% results: fixed and relatively flat manifolds, characteristic profile, untrained nets, mixup, curvature gap indicator
Our results show that learned representations span manifolds whose curvature is mostly fixed with relatively small values (on the order of $\expnumber{1}{-1}$), except for the output layer where curvature increases significantly (on the order of $1$). Moreover, this curvature profile was shared among several different convolutional architectures when considered as a function of the relative depth of the network. In particular, highly-curved data manifolds at the output layer have been observed in all cases, even in mixup-based models \cite{zhang2017mixup} which flatten intermediate manifolds more strongly in comparison to non mixup-based networks. In contrast, untrained models whose weights are randomly initialized presented a different curvature profile, yielding completely flat (i.e., zero curvature) manifolds towards the later layers. Further, our analysis suggests that estimates of dimensionality based on principal component analysis or more advanced methods need not reveal the actual characteristics of the curvature profile. Finally and similarly to indicators based on the intrinsic dimension~\cite{ansuini:NIPS:2019:intrinsic, birdal2021intrinsic}, we have found that the curvature gap in the last two layers of the network predicts its accuracy in that smaller gaps are associated with inferior performance, and larger gaps are related to more accurate models.

\section{Related Work}
\label{Sec:Related}

Geometric approaches commonly appear in learning-related tasks. In what follows, we narrow our discussion to manifold-aware learning and manifold-aware analysis works, and we refer the reader to surveys on geometric learning \cite{shuman2013emerging, bronstein2017geometric}. 

% manifold-aware learning
\paragraph{Manifold-aware learning.} Exploiting the intrinsic structure of data dates back to at least~\cite{belkin2004semi}, where the authors utilize the graph Laplacian to approximate the Laplace--Beltrami operator, which further allows to improve classification tools. More recently, several approaches that use geometric properties of the underlying manifold have been proposed. For instance, the intrinsic dimension (ID) was used to regularize the training of deep models, and it was proven to be effective in comparison to weight decay and dropout regularizers~\cite{zhu2018ldmnet}, as well as in the context of noisy inputs~\cite{ma2018dimensionality}. Another work~\cite{gong2019intrinsic} used the low dimension of image manifolds to construct a deep model. Focusing on symmetric manifolds, \citet{jensen2020manifold} propose a generative Gaussian process model which allows non-Euclidean inference. Similarly, \citet{goldt2020modeling} suggest a generative model that is amenable to analytic treatment if data is concentrated on a low-dimensional manifold. Other approaches aim for a flat latent manifold by penalizing the metric tensor \cite{chen2020learning}, and incorporating neighborhood penalty terms~\cite{lee2021neighborhood}. Additional approaches modify neural networks to account for metric information \cite{hoffer2015deep, karaletsos2015bayesian, gruffaz2021learning}. A recent work~\cite{chan2022redunet} showed that mapping distributions of real data, on multiple nonlinear submanifolds can improve robustness against label noise and data corruptions.

% manifold-aware analysis
\paragraph{Manifold-aware analysis.} \citet{basri2016efficient} explore the ability of deep networks to represent data that lies on a low-dimensional manifold. The intrinsic dimension of latent representations was used in~\cite{ma2018characterizing} to characterize adversarial subspaces, and to distinguish between learning styles with clean and noisy labels~\cite{ma2018dimensionality}. In~\cite{li2018measuring}, the authors employ random subspace training to approximate the ID, and to relate it to problem difficulty. Further, \citet{pope2021intrinsic} found that the ID is correlated with the number of natural image samples required for learning. Subsequently, \citet{kienitz2022effect} investigated the interplay between entanglement and ID, and their effect on the sample complexity. \citet{birdal2021intrinsic} harness the formalism of topological data analysis to estimate the ID, and they show it serves as an indicator for the generalization error. Perhaps closest in spirit to our study is the work~\cite{ansuini:NIPS:2019:intrinsic} where the ID is estimated on several popular vision deep architectures and benchmarks. Their results show that the intrinsic dimension follows a characteristic hunchback profile, and that the ID is negatively correlated with generalization error. Additionally, the authors speculate that latent representations in the final layer of neural networks are highly curved due to the large gap between the ID and the linear dimension (PC-ID) as measured by principal component analysis (PCA).

Beyond dimensionality, other works considered additional properties of the manifold. In~\cite{tosi2014metrics, arvanitidis2017latent}, the authors compute the Riemannian metric to obtain faithful latent interpolations. \citet{buchanan2020deep} studied how DNNs can separate two curves, representing the data manifolds of two separate classes, on the unit sphere.
The geometry in deep models with random weights was studied in~\cite{poole2016exponential}, where the authors find that curvature of decision boundaries flatten with depth, whereas data manifolds of e.g., a circular path increase their curvature along network layers. Similarly, \citet{kaul2019riemannian} also explore the curvature around the decision boundary, and they identify high curvature in transition regions. In contrast, \citet{fawzi2018empirical} identify that the decision boundary is mostly flat near data points.

The curvature of latent representations was estimated in~\cite{brahma2015deep} using deep belief networks \cite{hinton2006fast} with Swiss roll data and face images. One of the main conclusions was that the manifold flattens with depth. However, their curvature estimates were based on geodesic distances using the connectivity graph, and thus such estimates may be less reliable in settings of sparse and high-dimensional data manifolds. In contrast to~\cite{brahma2015deep}, it is shown in~\cite{shao2018riemannian} that manifolds learned with variational autoencoders for image data are almost flat. \citet{yu2018curvature} use curvature estimates to compare between two neural networks with respect to their fully connected layers. To stabilize computations, the authors propose to augment the data in the neighborhood of every sample. Overall, curvature characterization of latent representations related to deep convolutional models and benchmark datasets is still missing, and thus we focus the current research on this setting.

% \cite{moosavi:CVPR:2019:robustness} Use curvature as a regularizer of the loss surface with respect to inputs which enforces the trained neural network be more more linear. In addition, they provide theoretical evidence that show a relation between small curvature and  robustness of  of adversarially trained models .

% Papers that claim their method flattens the objective manifold without
% providing theoretical or empirical evidence
% The work of \cite{he:ICM:2004:learning} propose a method that mapping function from feature space to high-level semantic space which claimed to flat. 
\section{Background and Method}
\label{Sec:Method}

% ID under/over estimation for ID<20  => TOOO: something similar in RC?
% ID is scale-invariant (check via dataset decimation), tried on data with known ID embedded in 100,000 dim

% high-level overview: dataset, train model, collect latent representations, compute curvature quantities (neighborhood + hessian approx.), analysis
Given a dataset (e.g., CIFAR-10) and an architecture (e.g., ResNet18), we train the model on the data, and we collect its latent representations along the layers of the model for the train and test sets. Curvature information is estimated for the latent codes, and we perform our analysis on a single curvature quantity, typically the mean absolute value of principal curvatures (see the discussion in App.~\ref{App:cmp_curv_metrics}), and on the distribution of principal curvatures. In what follows, we briefly describe the extraction of latent codes and the curvature estimation procedure.

% improve density of data: curvature is 2nd derivative of manifold, high-dim. sparse, curse of dim., generate neighborhood, pass through network, first layer irrelevant, other representations are close enough
\paragraph{Data density.} In contrast to the intrinsic dimension which is a global feature of the manifold (for connected manifolds), curvature information is a \emph{local property}~\cite{lee2006riemannian}. Additionally, curvatures are based on \emph{second-order derivatives} of the manifold. Thus, our investigation makes the implicit assumption that data is sufficiently dense for computing curvatures. However, datasets that frequently appear in machine learning, e.g., CIFAR-10, are high-dimensional and sparse, and thus computing local differentiable quantities on such data is extremely challenging. 

\begin{figure}[!t]
    \centering
    \begin{overpic}[width=1\linewidth]{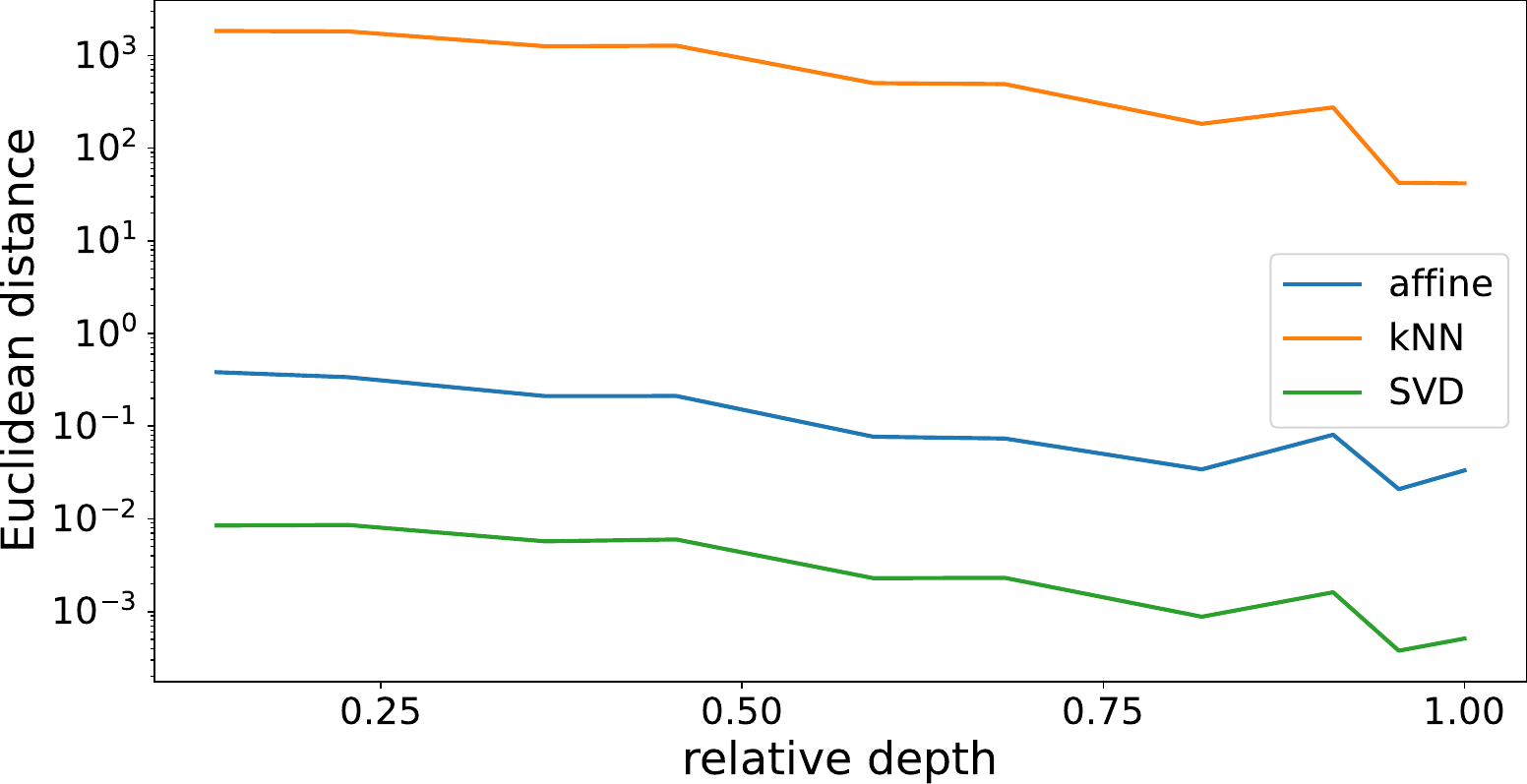}
       % \put(0, 30){A} \put(55, 30){B}
    \end{overpic}
    \vspace{-3mm}
    \caption{We plot the average Euclidean distance of neighborhoods points to their center with respect to the network relative depth. SVD yields denser neighborhoods, and thus we use it in our study.}
    \label{Fig:neigh_dists}
\end{figure}

The above characteristics of typical machine learning data require a large number of close points for creating a stable neighborhood. To this end, a commonly-used tool is k-Nearest-Neighbours (KNN). Unfortunately, this method depends on the closeness of points in the dataset, and thus it may generate non-local and spare neighborhoods where ``neighbors'' are effectively far in a Euclidean sense. Another common method is to use domain-specific augmentations. For instance, applying image transformations such as rotation and scaling, based on the assumption that natural images are invariant to these geometric transforms. However, geometric manipulations explore only a particular aspect of the data manifold, while potentially ignoring other parts. An effective domain-agnostic approach computes the Singular Value Decomposition (SVD) per data point, and it generates a close neighborhood by filtering out small amounts of noise in the data. The approach is well motivated from a differential geometry viewpoint as it is closely related to computing a first-order approximation of the manifold at a point, and sampling the point neighborhood. We provide a detailed comparison between the above three methods in App.~\ref{App:data_density}. In addition, we show in Fig.~\ref{Fig:neigh_dists} the average distance to the center of points generated with affine transformations, kNN and SVD, across the network layers. Notably, SVD provides denser neighborhoods, and thus we use this approach in our study as detailed below.

\paragraph{Neighborhood generation.} To improve the local density of image samples, we use the same procedure as in~\cite{yu2018curvature} to generate artificial new samples by reducing the ``noise'' levels of the original data. Specifically, given an image $I\in \mathbb{R}^{m \times n \times c}$, we denote by $I_{j}\in \mathbb{R}^{m \times n}$ the matrix at channel $j$. Let $I_{j}=U \Sigma V^T$ be its SVD, where $U\in \mathbb{R}^{m \times m}$, $V\in \mathbb{R}^{n \times n}$ and $\Sigma \in \mathbb{R}^{m \times n}$ a rectangular diagonal matrix with singular values $\left\{\sigma_{1}, \sigma_{1}, \cdots, \sigma_{r} \right\} $ on the diagonal in descending order such that $r$ is the rank of $I_{j}$. We define $\Sigma'$ as the result of zeroing a subset of singular values in $\Sigma$, allowing to create a new close image $I_{j}'=U \Sigma' V^T$. This process is performed along all three R, G, B layers. In our experiments we zeroed all combinations of the ten smallest singular values, generating $1024$ new images. %This process can be thought of as removing a small portion of noise.

% extraction process: image+neigh passed through net, only certain layers (as in ID), no input layer, TODO: support robustness claim
\paragraph{Latent representations.} Given an image $I$ of the data of interest, we generate its neighborhood samples using the procedure above, denoted by $\{ I'(i) \}$ for $i=1,\dots, 1024$. We pass the original image and its neighborhood through the network, and our curvature analysis is performed separately on every such batch. Importantly, passing the input batch of the image and its neighborhood through the nonlinear transformations of the network yields an approximation of a local patch on the manifold, allowing for robust curvature computations. In practice, we extract the latent codes of a subset of layers, similarly to~\cite{ansuini:NIPS:2019:intrinsic}. For instance, in the experiments with ResNets we use the latent codes after every ResNet block and the average pooling before the output. We note that our analyses includes curvature information in the input layer, even though it is shared across different architectures for the same dataset. 

% curvature estimation: multiple approaches for curvature quantities, RC estimation is a linear regression problem,
\paragraph{Curvature estimation.} There are multiple approaches to estimate curvature quantities of data representations, see e.g., \cite{brahma2015deep, shao2018riemannian}. We decided to use the algorithm presented in \cite{li2018curvature} and named Curvature Aware Manifold Learning (CAML) since it is backed by theory and is relatively efficient. CAML requires the neighborhood of a sample, and an estimate of the unknown ID. The ID is computed using the TwoNN algorithm~\cite{facco:SR:2017:twonn} on the original dataset (without augmentation) per layer, similarly to~\cite{ansuini:NIPS:2019:intrinsic}. 

Let $Y=\{y_1, y_2, \cdots, y_N \}\subset\mathbb{R}^{D}$ be the data on which we want to estimate the curvature. We assume that the data lies on a $d$-dimensional manifold $\mathcal{M}$ embedded in $\mathbb{R}^{D}$ where $d$ is much smaller than $D$, thus, $\mathcal{M}$ can be viewed as a sub-manifold of $\mathbb{R}^{D}$. The key idea behind CAML is to compute a second-order local approximation of the embedding map,
\begin{equation}
    \label{EQN:embmap}
    f: \mathbb{R}^d \rightarrow \mathbb{R}^D \quad, y_i = f(x_i) + \epsilon_{i} \ , \quad i=1,\dots,N \ ,
\end{equation}
where $X=\{x_1, x_2, \cdots, x_N\}\subset\mathbb{R}^{d}$ are low-dimensional representations of $Y$, and $\{\epsilon_1, \epsilon_2, \cdots \epsilon_N\}$ are the related noises. In the context of this paper, the embedding map $f$ is the transformation that maps the low-dimensional image representations to a pixel-wise form that might hold redundant information.

To estimate curvature information at a point $y_i \in Y$, we define its neighborhood via the procedure described above, yielding a set of close points $\{ y_{i_1}, \dots, y_{i_K} \}$ where $K$ is the number of neighbors. We use this set and the point $y_i$ to construct via SVD a local natural orthonormal coordinate frame $\left\{\frac{\partial}{\partial x^1}, \cdots, \frac{\partial}{\partial x^d}, \frac{\partial}{\partial y^1}, \cdots, \frac{\partial}{\partial y^{D-d}}\right\}$, composed of a basis for the tangent space (first $d$ elements), and a basis for the normal space. We denote by $x_i$ and $u_{i_j}$ the projection of $y_i$ and $y_{i_j}$ for $j=1, \dots, K$ to the tangent space spanned by $\partial/\partial x^1, \dots, \partial/\partial x^d$, respectively. Importantly, the neighborhood of $y_i$ must be of rank $r>d$, otherwise, SVD can not encode the normal component at $x_i$, yielding poor approximations of $f$ at $x_i$. Thus, we verify that $\{y_{i_1}, \dots, y_{i_K} \}$ is of rank $d+1$ or more.

The map $f$ can then be re-formulated in the latter coordinate frame as $f(x^1,\dots,x^d) = [x^1, \dots, x^d, f^1, \dots, f^{D-d}]$. The second-order Taylor expansion of $f^\alpha$ at $u_{i_j}$ with respect to $x_i$ and up to $\mathcal{O}(|u_{i_j}|_2^2)$ error is given by
\begin{equation}
    \label{EQN:maptaylor}
    f^\alpha(u_{i_j}) \approx f^\alpha(x_i) + \Delta _{x_i}^T \nabla f^\alpha +\frac{1}{2} \Delta _{x_i}^T H^\alpha \Delta _{x_i} \ ,
\end{equation}
where $\alpha=1,\dots,D-d$, $\Delta _{x_i}=(u_{i_j}-x_i)$ and $u_{i_j}$ is a point in the neighborhood of $x_i$. The gradient of $f^\alpha$ is denoted by $\nabla f^\alpha$, and $H^\alpha = \left( \frac{\partial^2 f^\alpha}{\partial x^i \partial x^j} \right)$ is its Hessian. Given a neighborhood $\{ y_{i_1}, \dots, y_{i_K}\}$ of $y_i$, and their corresponding tangent representations $\{ u_{i_j} \}$, we can use Eq.~\ref{EQN:maptaylor} to form a system of linear equations, as we detail in App.~\ref{App:taylor_linsolve}. The principal curvatures are the eigenvalues of $H^\alpha$, and thus estimating curvature information is reduced to a linear regression problem followed by an eigendecomposition. Each Hessian has $d$ eigenvalues, therefore each sample will have $\left(D-d \right) \times d$ principal curvatures. Finally, we note that one can potentially also compute the Riemannian curvature tensor using the principal curvatures~\cite{yu2018curvature}. However, the latter tensor has an order of $d^4$ elements, and thus its evaluation demands high computational resources. Further, as the Riemannian curvature tensor is fully determined by the principal curvatures, we base our analysis on the eigenvalues of the Hessian. To evaluate the curvature of manifolds, we estimate the mean absolute principal curvature (MAPC) which is given by the mean of the absolute values of eigenvalues of the estimated Hessian matrices.

\begin{figure*}[h]
    \centering
    \begin{overpic}[width=1\linewidth, ,height=5cm]{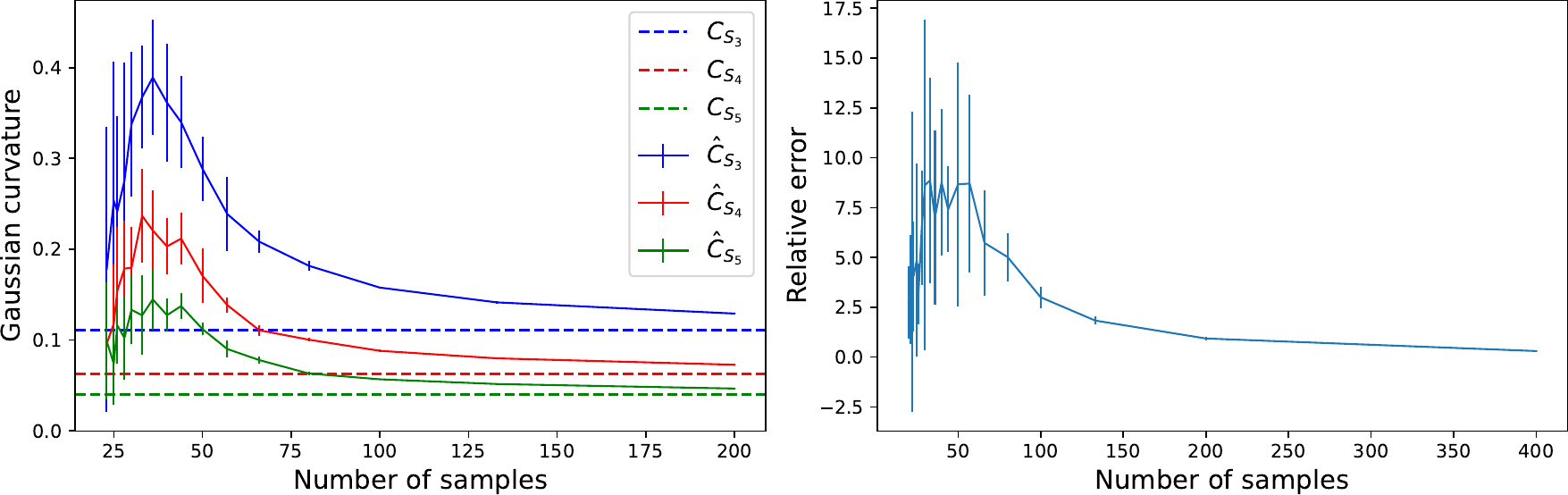}
        \put(0, 28) {A} \put(50, 28) {B}
    \end{overpic}
    \caption{\textbf{Estimation of the curvature of the surface of a sphere and the surface of an ellipsoid using the CAML algorithm.} \textbf{A)} Estimation of the curvature of spheres with different radii where $C_{S_i}$ is the Gaussian curvature of a sphere with radius $i$ and $\hat{C}_{S_i}$ is the estimated curvature using the CAML algorithm. \textbf{B)} The relative error between the estimation and the analytic value of the Gaussian curvature of an ellipsoid.}
    \label{Fig:sphere_ellipsoid}
\end{figure*}

\paragraph{CAML evaluation.} The CAML algorithm~\cite{li2018curvature} was published without implementation and had a few minor issues. To test our implementation we focused on two 2-dimensional manifolds: spheres and ellipsoids. The Gaussian curvature at a point $p$ of these manifolds has a closed-form formulation defined as the product of the principal values at $p$. For instance, a sphere of radius $r$ has a Gaussian curvature of $\frac{1}{r^{2}}$ everywhere. Fig.~\ref{Fig:sphere_ellipsoid} shows the Gaussian curvature estimation of three different spheres using the CAML algorithm. In contrast to the surface of a sphere, the points on the surface of an ellipsoid have different Gaussian curvature values, given by the following equation:
\begin{equation} \label{Eqn:ellipsoid_crv}
K(x, y, z)=\left (a^2 b^2 c^2\left(\frac{x^2}{a^4}+\frac{y^2}{b^4}+\frac{z^2}{c^4}\right)^2 \right)^{-1} \ ,
\end{equation}
where $a,b$, and $c$ are the parameters that define the ellipsoid:
\begin{equation} \label{Eqn:ellipsoid_eqn}
\left(\frac{x}{a}\right)^2+\left(\frac{y}{b}\right)^2+\left(\frac{z}{c}\right)^2=1 \ .
\end{equation}
It is noticeable that the relative error quickly decreases to zero as samples increase, meaning that the CAML algorithm is able to estimate curvatures of object manifolds reliably.

\section{Results}
\label{Sec:Results}

\begin{figure*}[h]
    \centering
    \begin{overpic}[width=1\linewidth]{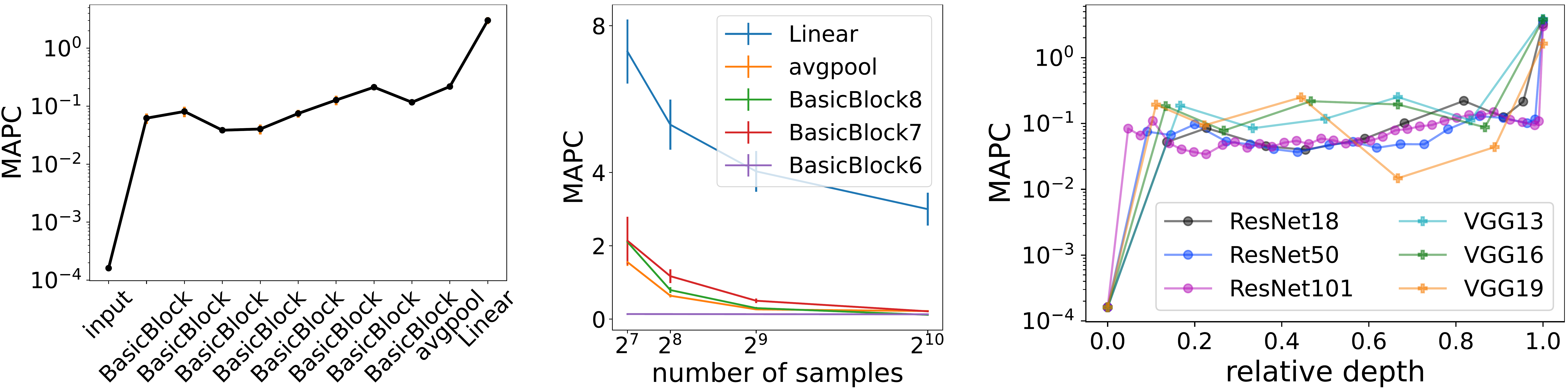}
        \put(1, 23){A} \put(34, 23){B} \put(65, 23){C}
    \end{overpic}
    \caption{\textbf{Mean absolute principal curvature along layers of deep convolutional networks.} \textbf{A)} MAPC and standard deviation as measured for ten seeds using ResNet18 on CIFAR-10 train set. \textbf{B)} Repeated evaluation of MAPC on sub-sampled neighborhoods converges for $1024$ elements. \textbf{C)} MAPC graphs for VGG and ResNet families as a function of the model's relative depth, presenting a characteristic step-like shape in all cases.}
    \label{Fig:common_mapc}
\end{figure*}

\subsection{Data manifolds feature a common curvature profile}
\label{Subsec:mapc}

% MAPC exhibits a characteristic profile
We begin our analysis with an empirical evaluation of the curvature of latent representations along the layers of a ResNet18 network~\citep{he2016deep}, trained on the CIFAR-10 dataset~\citep{krizhevsky2009learning}. For a selected subset of layers, we estimate the mean absolute principal curvature (MAPC) as described in Sec.~\ref{Sec:Method}. We repeat MAPC evaluation for ten models initialized with random seeds, and we show standard deviation per layer in orange. In Fig.~\ref{Fig:common_mapc} we observe that MAPC is generally \emph{increasing} with depth, demonstrating a large variation of almost four orders of magnitude: $\text{MAPC(Input)} = \expnumber{1.6}{-4}$ and $\text{MAPC(Linear)} = 3.0$, see Fig.~\ref{Fig:common_mapc}A. Additionally, sharp increases in curvature occur during the transition between the input and the output of the following layer (BasicBlock) as well as between the penultimate to last layers (avgpool to Linear). Notably, MAPC is relatively \emph{fixed} for a majority of network layers.

% robustness of algorithm
Our curvature estimates depend directly on the neighborhood around each data point, see Sec.~\ref{Sec:Method}. Specifically, sparse and noisy neighborhoods may lead to poor estimates of curvature. We evaluate the robustness of our MAPC computations by evaluating CAML on a repeated sub-sampling of the neighborhood. We observe an overall stable behavior for MAPC values along the last five layers of ResNet18, see Fig.~\ref{Fig:common_mapc}B. In particular, MAPC values stabilize in terms of standard deviation when the number of samples per neighborhood reached $1024$ elements, and thus we collect $1024$ samples in all of our experiments.

% relative depth vs. MAPC
We further our exploration by investigating whether the characteristic ``step-like'' shape of MAPC shown in Fig.~\ref{Fig:common_mapc}A is shared across multiple networks and datasets. We repeated the above analysis for three variants of a VGG architecture (VGG13, VGG16, VGG19) and three variants of a ResNet architecture (ResNet18, ResNet50, ResNet101) trained on CIFAR-10 and CIFAR-100 datasets, for a total of $12$ different models. We show in Fig.~\ref{Fig:common_mapc}C six MAPC profiles obtained for CIFAR-10 and plotted with respect to the \emph{relative depth} of the network. Similarly to \cite{raghu2017svcca}, we define the relative depth as the absolute depth of the layer divided by the total number of layers, not counting batch normalizations. Remarkably, the MAPC profiles reveal a common step-like shape, despite the large variation in the underlying models in terms of overall structure, number of layers, and regularization methods. Beyond their shared behavior, all MAPC graphs attain similar absolute values across network layers, overlapping particularly in the last layer. See a qualitatively similar plot for CIFAR-10 test set, and CIFAR-100 train set in Fig.~\ref{Fig:common_mapc_app}.

Our results identify that curvature of data manifolds admits a particular trend including three phases: an initial increase, followed by a long phase of a plateau, and ending with an abrupt final increase. These results are consistent with theoretical studies~\citep{cohen:NATURE:2020:separability}, and empirical explorations on neural networks with random weights~\citep{poole2016exponential}. Particularly relevant are the findings in~\citep{ansuini:NIPS:2019:intrinsic}, showing low values of intrinsic dimension (ID) in the last layer of deep convolutional networks, and a large gap between the ID and its linear estimation (PC-ID). The authors propose an indicator for the generalization of the model to unseen data based on the ID values in the last hidden layer, and additionally, they related the gap between PC-ID and ID to the curvature of the data manifold. Motivated by their results and analysis, we suggest a new curvature-based generalization indicator (\ref{Subsec:pc_gen_err}), and we study the relation between dimensionality and curvature (\ref{Subsec:id_pc}).

% \begin{figure*}[ht]
%     \centering
%     \begin{overpic}[width=1\linewidth]{figures/acc_vs_mapc_ngap}
%         \put(0, 30){A} \put(55.5, 30){B}
%         \put(19, 30){CIFAR-10} \put(18, 14.5){CIFAR-100}
%     \end{overpic}
%     \caption{\textbf{Normalized MAPC gap is correlated with accuracy.} \textbf{A)} Normalized MAPC gap with respect to model accuracy for six different networks on CIFAR-10 and CIFAR-100. \textbf{B)} Normalized MAPC gap with respect to accuracy for six different networks on subsets of CIFAR-100, see text.}
%     \label{Fig:acc_vs_mapcs_ngap}
% \end{figure*}

\begin{figure*}[ht]
    \centering
    \begin{overpic}[width=1\linewidth]{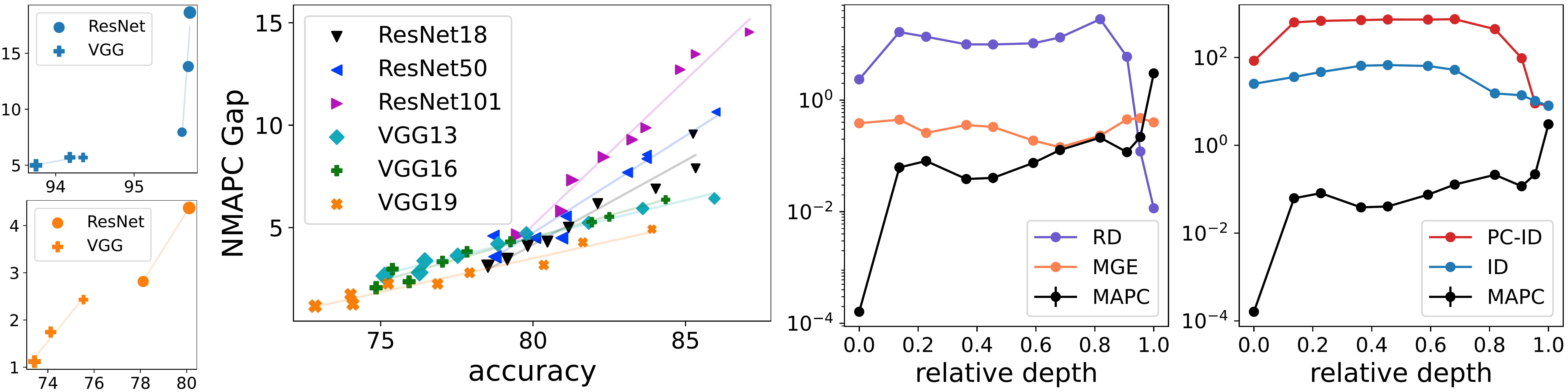}
        \put(0, 23){A} \put(14, 23){B}
        % \put(19, 30){CIFAR-10} \put(18, 14.5){CIFAR-100}
        \put(51, 23){C} \put(76, 23){D}
    \end{overpic}
    \caption{\textbf{NMAPC gap is correlated with accuracy.} \textbf{A)} Normalized MAPC gap with respect to model accuracy for six different networks on CIFAR-10 (top) and CIFAR-100 (bottom). \textbf{B)} Normalized MAPC gap with respect to accuracy for six different networks on subsets of CIFAR-100, see text. \textbf{Comparison of dimensionality and curvature.} \textbf{C)} Relative difference between linear dimension and intrinsic dimension, and maximum gap in eigenvalues of the covariance matrix are compared with MAPC along the relative depth of ResNet18 trained on CIFAR-10. \textbf{D)} PC-ID, ID and MAPC for the same network.}
    \label{Fig:mapc_id_gap}
\end{figure*}
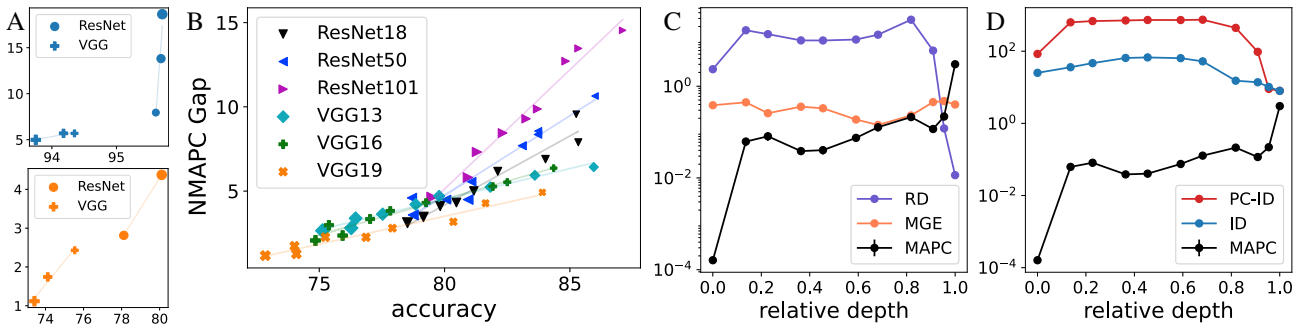

% MAPC predicts generalization gap? performance? acc./top 5% vs. gap
\subsection{Curvature gap in final layers is correlated with model performance}
\label{Subsec:pc_gen_err}

Our empirical results regarding the curvature profiles for CIFAR-10 and CIFAR-100 (Figs.~\ref{Fig:common_mapc}, \ref{Fig:common_mapc_app}) indicate that MAPC values are higher for CIFAR-10. Moreover, the difference between curvatures in the last two layers of the network, termed MAPC gap from now on, are noticeably smaller for CIFAR-100. In addition, curvature values vary across different models trained on the same dataset. These differences led us to investigate whether the MAPC gap is correlated with the performance of CNNs across architectures and datasets. Specifically, we consider the \emph{normalized MAPC (NMAPC) gap} defined as the MAPC gap divided by the average of MAPC across layers, and we compare it against the accuracy of the network. We evaluate the normalized gap on the train sets of CIFAR-10 and CIFAR-100 for the ResNet and VGG families. Each data point corresponds to one of the six models, where the size of the marker represents the network size, e.g., smallest marker for ResNet18 and largest marker for ResNet101 (Fig.~\ref{Fig:mapc_id_gap}A). We observe a remarkable correspondence between model performance and the NMAPC gap, also emphasized by the additional linear fit graphs per network family. These linear graphs show a consistent trend per family with respect to the difference in gap in relation to difference in accuracy. 

To further investigate the correlation between the NMAPC gap and model performance, we perform the following experiment. We divide the CIFAR-100 dataset which contains a hundred different classes $c_{1}, c_{2}, \dots, c_{100}$ to ten subsets $i\in\{1, \dots, 10\}$ such that subset $i$ contains samples from classes $c_{1}$ to $c_{10i}$. We trained all six networks on all subsets, and we computed the NMAPC gap and compared it with model performance (Fig.~\ref{Fig:mapc_id_gap}B). To improve visibility, we use a different color for every network. Per architecture, each data point corresponds to one of the subsets, where its size represents the size of the subset, e.g., largest markers for the full CIFAR-100 dataset. Similarly to Fig.~\ref{Fig:mapc_id_gap}A, we augment the plot with linear fit graphs per architecture. In \emph{all} models and subsets, we find a remarkable correlation between the NMAPC gap and accuracy value. We emphasize that similarly to the ID indicator~\citep{ansuini:NIPS:2019:intrinsic}, the NMAPC gap can be employed without accessing the test set.

\subsection{Dimensionality and curvature of data manifolds need not be correlated}
\label{Subsec:id_pc}

% gap between PC-ID and ID correlated to MAPC? ID and MAPC correlated?
% relative comparison, std/mean?
Our third analysis explores the relation between dimensionality and curvature of the data manifold. Existing work on data representations assumes there is a correlation in the flatness of the manifold with respect to dimensionality measures~\citep{verma2019manifold, ansuini:NIPS:2019:intrinsic}. On the other hand, analytic examples in geometry such as minimal surfaces where the principal curvatures are equal and opposite at every point~\citep{do2016differential}, tell us that dimensionality and curvature need not be related. Motivated by these considerations, we ask: how does the dimension correspond to curvature along the network's layers? 

To address this question, we extracted the latent representations of a ResNet18 network trained on the CIFAR-10 dataset, and we computed the linear dimension (PC-ID), intrinsic dimension (ID), and mean absolute principal curvature (MAPC). Following \cite{ansuini:NIPS:2019:intrinsic}, PC-ID is defined to be the number of principal components that describe $90\%$ of the variance in the data, and ID is computed using TwoNN~\citep{facco:SR:2017:twonn}. We focus on the \emph{relative} absolute difference between PC-ID and ID, i.e., $\text{RD} := |\text{PC-ID} - \text{ID}| / \text{ID}$, as a proxy for inferring curvature features, see Fig.~\ref{Fig:mapc_id_gap}C. In comparison to the MAPC profile (black), we found no correlation with the relative difference (purple). For instance, RD is high in the first two layers, whereas MAPC is low in the first layer, and then it increases significantly in the second layer. Notably, RD and MAPC admit a weak inverse correlation toward the last three layers of the model. Additionally, we estimate the maximum gap in the eigenvalues of the normalized covariance matrix given by $\text{MGE} := \max_j (\bar{\lambda}_j - \bar{\lambda}_{j+1})$, where $\bar{\lambda}_j$ are the eigenvalues scaled to the range $[0, 1]$. Similarly to the relative difference graph (RD), the maximum gap in eigenvalues (MGE) colored in orange generally does not correspond to MAPC. In particular, MGE in the first and last layers are close in value, whereas MAPC exhibits a difference of four orders of magnitude in those same layers. We also plot PC-ID, ID, and MAPC for the same network in Fig.~\ref{Fig:mapc_id_gap}D, showing the non-relative dimension estimates and MAPC.

% TODO: re-generate this plot
\begin{figure}[!b]
    \centering
    \includegraphics[width=.7\linewidth]{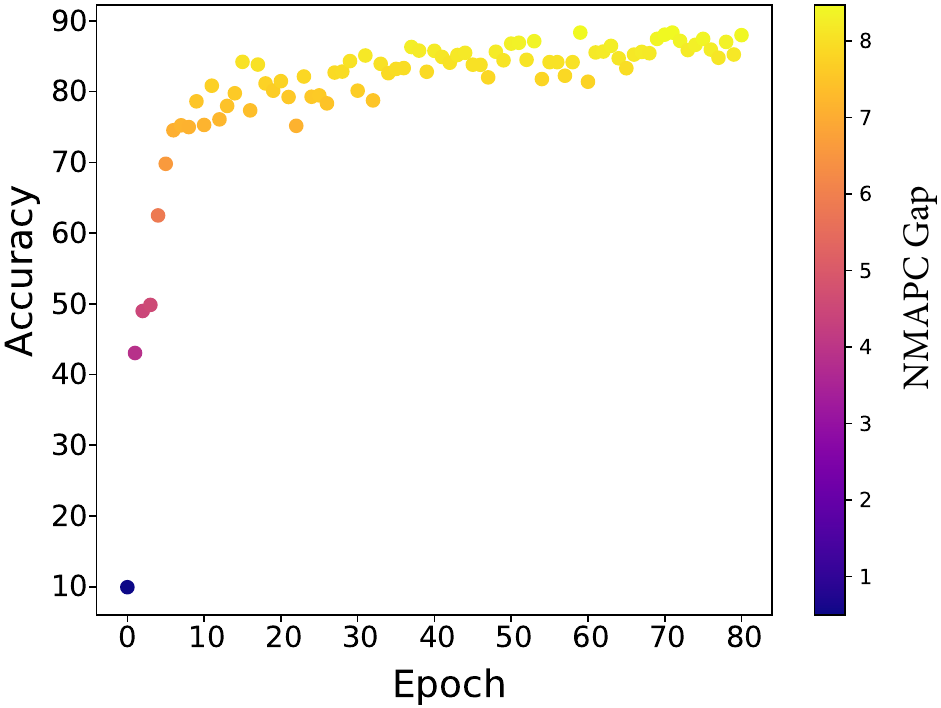}
    \caption{\textbf{Training dynamics of the NMAPC gap on ResNet18 and CIFAR-10.} The plot shows how the accuracy changes during training, colored by the normalized curvature gap.}
    \label{Fig:train_dynamics}
\end{figure}

% Training dynamics
\subsection{Training dynamics} 
\label{SSec:train_dynamics}

Due to the correspondence between model performance and the NMAPC gap, we were interested to see if the training process of the network increases the mentioned gap. We trained a Resnet18 network with CIFAR-10 and observed how the gap changes. We hypothesized that the gap will increase as the network training converges. Remarkably, we indeed find that the NMAPC gap is highly correlated with the behavior of the network during training (Fig.~\ref{Fig:train_dynamics}). Each dot in the plot represents a different snapshot of the model during training, and it is positioned with respect to its accuracy on the test set as a function of the epoch. The points are colored by their NMAPC gap (see color bar on the right). Overall, we observe that during training the accuracy increase in conjunction with the gap, meaning that the network favors a large gap to increase its performance.

\begin{figure*}[t]
    \centering
    \begin{overpic}[width=1\linewidth]{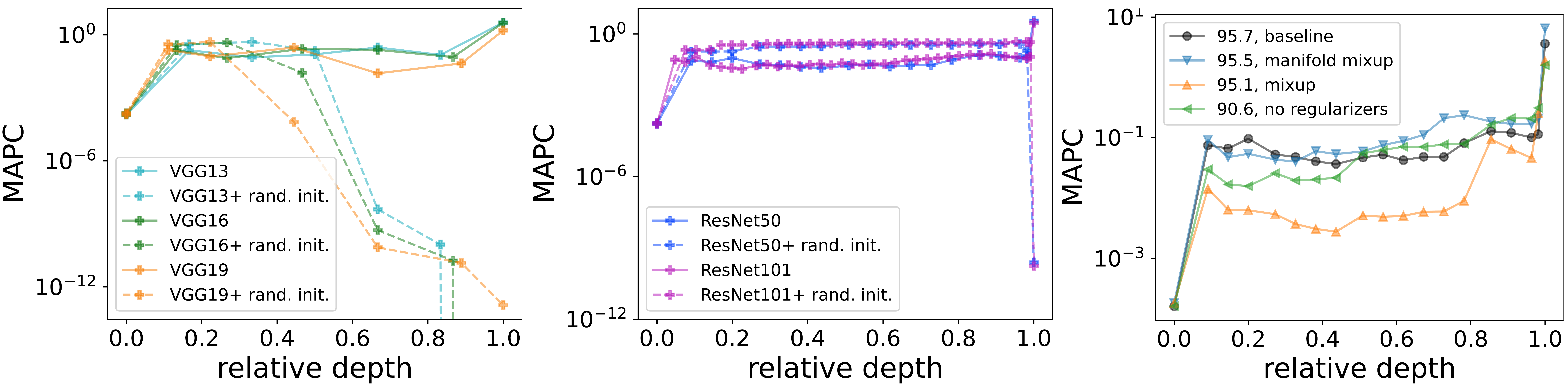}
        \put(1, 23){A} \put(35, 23){B} \put(68, 23){C}
    \end{overpic}
    \caption{\textbf{Comparison of MAPC profiles for baseline models with untrained networks and regularized networks.} \textbf{A)} MAPC graphs for VGG neural architectures before and after training. \textbf{B)} MAPC graphs for ResNet neural architectures before and after training. \textbf{C)} MAPC graphs for ResNet50 networks trained with regularizers such as weight decay, learning rate scheduling, and mixup.}
    \label{Fig:mapc_rand_n_reg}
\end{figure*}

% random initialization (non-trained), random labels??
\subsection{Untrained networks exhibit a different profile}
\label{Subsec:untrained}

We also computed curvature estimates of data representations along the layers of VGG13, VGG16, and VGG19 for randomly initialized networks. In comparison to MAPC profiles of trained networks (solid lines in \ref{Fig:mapc_rand_n_reg}A), untrained models demonstrate significantly different trends (dashed lines in \ref{Fig:mapc_rand_n_reg}A). While curvature profiles of randomly initialized models and trained networks approximately match up until half of the network depth, there is a sharp decrease in MAPC for untrained models in the second half. Importantly, MAPC values present a similar increase in the first layers for all models, whereas, in the final layers of untrained networks MAPCs are essentially zero. We also note that the decrease in curvature is steeper for larger networks---the orange line (VGG19) is lower than the green line (VGG16), which in turn, is lower than the cyan line (VGG13), except for the final layer. ResNet architectures also present a different profile, with constant MAPC along layers and a sharp decrease at the final layer, Fig.~\ref{Fig:mapc_rand_n_reg}B. These results indicate that MAPC profiles of deep convolutional neural networks initially depend on the structure of the model, however, the behavior in the last layers is a direct result of training.

% curvature wrt regularization
\subsection{The effect of standard regularizers on curvature}

Regularization is a common practice for modern neural models which are often overparameterized, i.e., the amount of trainable weights is significantly larger than the amount of available train data~\citep{allen2019learning}. Beyond limiting the parameter space to preferable minimizers, and leading to better generalization properties, certain regularization techniques may affect additional features of the task. For instance, mixup-based methods which augment train data with convex combinations of the inputs and labels \citep{zhang2017mixup} are associated with the flattening of the data manifold~\citep{verma2019manifold}. In their context, flattening means that significant variance directions on the data manifold are reduced. Our curvature estimation framework motivates us to further ask: how do typical regularizers affect curvature statistics of convolutional neural networks?

In the following experiment we investigate this aspect with the baseline model ResNet50 used throughout the paper. The ResNet50 net is trained with weight decay of $\expnumber{5}{-4}$ and cosine annealing learning rate scheduling. Additionally, we also train this network with no regularizers, and with manifold mixup and mixup (and no other regularization). We find that all four models demonstrate a step-like profile (Fig.~\ref{Fig:mapc_rand_n_reg}C) consistent with our results (Fig.~\ref{Fig:common_mapc}). In particular, the plateau regime and high final MAPC were observed across all models. Notably, while the networks attained different curvature values in the last layer, the normalized MAPC gap (Fig.~\ref{Fig:mapc_id_gap}) distinguishes between the models, and it is correlated with their performance. Namely, we obtain $14, 14, 13, 8$ normalized MAPC gaps for the baseline, manifold mixup, mixup, and no regularization networks, respectively (see their test set accuracy in the legend of Fig.~\ref{Fig:mapc_rand_n_reg}C). As per flattening of the data manifold, we note that manifold mixup admits an MAPC profile close in values to our baseline model, whereas mixup shows a significant reduction in curvature (an order of magnitude along most layers in comparison to baseline). Remarkably, mixup does seem to flatten data representations in intermediate layers although it only alters the training samples. In contrast, manifold mixup which manipulates latent codes in a similar fashion to mixup, does not seem to affect MAPC values much. Further, these results reinforce our findings above that high curvature in the last layer, or more precisely, high normalized MAPC gap, is fundamental to the success of the learning model.

\begin{figure}[!hb]
    \centering
    \begin{overpic}[width=1\linewidth]{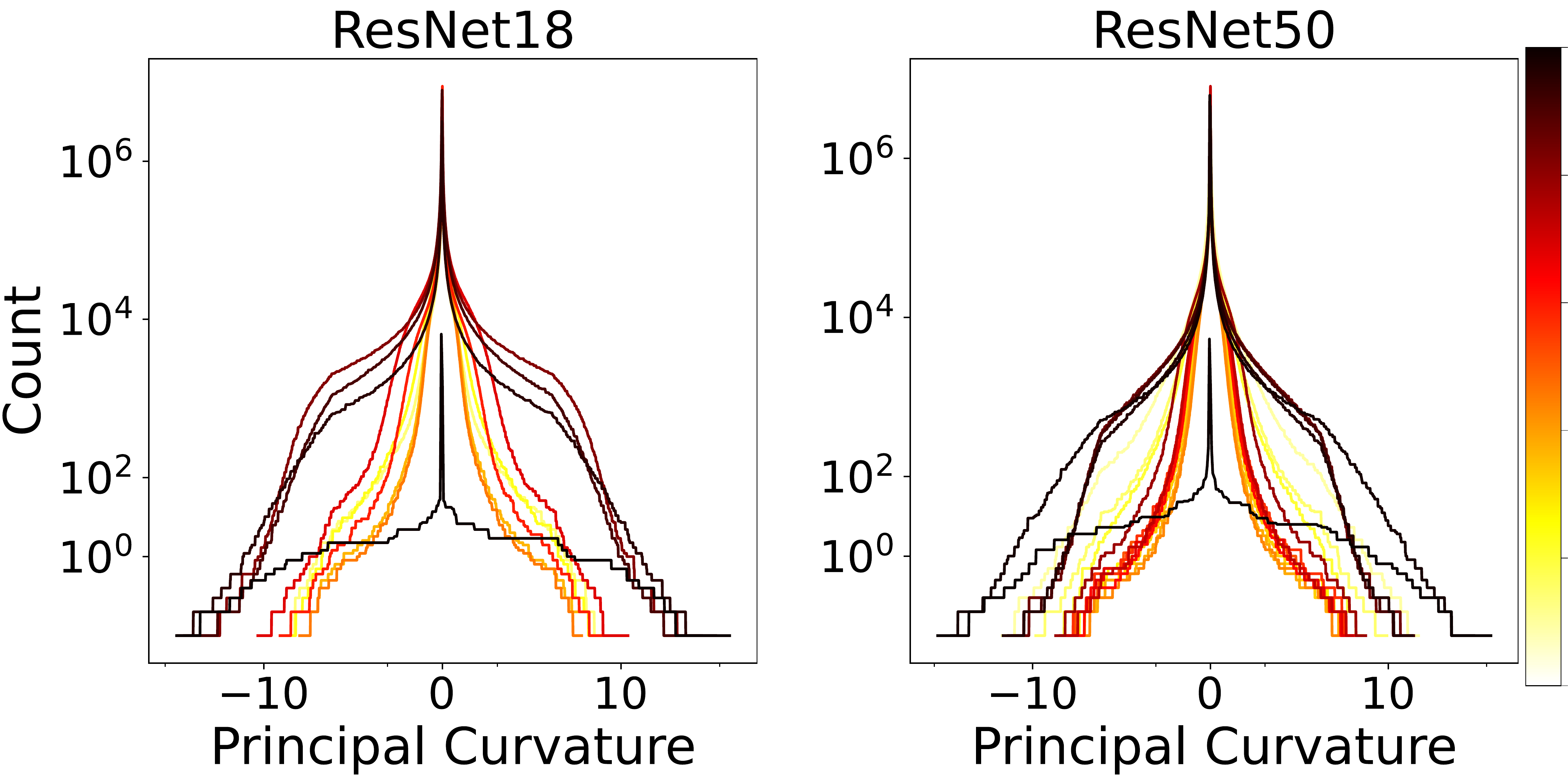}
    \end{overpic}
    \vspace{-3mm}
    \caption{\textbf{Distribution of principal curvatures for ResNet models.} Each plot shows the histogram profiles of principal curvatures per layer, colored by their relative depth.}
    \label{Fig:pc_distrib}
\end{figure}

% Geometry of data manifold? hist of PC per layer
\subsection{Distribution of principal curvatures}

In our analysis above, we focused on a single estimate of curvature for the entire manifold based on the average absolute value of principal curvatures (i.e., eigenvalues of the Hessian). However, we recall that curvature is a local property for each point of the manifold, and thus curvature variability should also be investigated. Here, we inspect the distribution of principal curvatures for all points at every layer. We estimate $(D-d)$ Hessian matrices for each of the $1000$ input images for the ResNet family (ResNet18, ResNet50), resulting in $d(D-d)\cdot 1000$ principal curvatures per layer. To analyze this massive amount of information, we compute a histogram per layer, and we plot them overlayed and differentiated by colors according to their relative depth (Fig.~\ref{Fig:pc_distrib}). For example, light curves are related to the initial layers of the network, whereas dark curves are associated with final network layers. Notably, we observe similar histogram profiles for the majority of intermediate layers (yellow to red curves) across all architectures both in terms of histogram shape and spread of values. Subsequent layers (dark red to brown) present wider distributions, achieving curvature values on the range of $10^2$. Indeed, we observed a mild increase in MAPC toward the last layer of the network (Fig.~\ref{Fig:common_mapc}C). The final layer shows that relatively more points attained non-zero curvatures, yielding a histogram profile with a wider base. This result confirms the sharp increase in MAPC of the last layer of CNNs as shown in (Fig.~\ref{Fig:common_mapc}C).

\section{Discussion}
\label{Sec:Discussion}

% image classification, ML and neuroscience, manifold untangling, no untangling measures, untangling and flattening, decision boundary vs. data manifold
Image classification is a fundamental task which is heavily studied in neuroscience and machine learning. Common wisdom on this problem suggest that \emph{untangling} of manifolds occurs throughout image processing by our vision system and brain~\citep{dicarlo2007untangling}, and by deep convolutional neural networks~\citep{bengio2013representation}. While manifold untangling is commonly perceived as ``simpler separability'' between class objects (often termed linear separability), defining formal measures of untangling is still an active research topic~\citep{chung2018classification}. Manifold untangling is typically mentioned alongside \emph{flattening} of the data manifold, a notion related to curvature and to Riemannian geometry. A recent work on this topic distinguishes between the curvature of the decision boundary, and the curvature of the data manifold~\citep{poole2016exponential}, identifying a flattening of the decision boundary with depth and an \emph{opposite} behavior of the data manifold, on deep neural networks with random weights. Additional theoretical and empirical studies provide a mixed picture on this topic, where some works observe flat decision boundaries~\citep{fawzi2018empirical}, and others report highly-curved transition regions~\citep{kaul2019riemannian}. Further, \cite{brahma2015deep} describe the flattening of data manifolds with depth, whereas \cite{shao2018riemannian} essentially observe flat representations. This large variance in results may be attributed to the large variety of different architectures and datasets considered in these works. In this context, our study is the first to investigate systematically how the curvature of latent representations change in common state-of-the-art deep convolutional neural networks used for image classification. 

Complementary to existing work on geometric properties of data representations involving their intrinsic dimension \citep{ansuini:NIPS:2019:intrinsic, birdal2021intrinsic}, and density evolution \citep{doimo2020hierarchical}, our study characterizes the \emph{curvature profile} of latent manifolds. The aggregated knowledge arising from prior works on convolutional networks indicate that the intrinsic dimension presents a rapid increase over the first layers, and then it progressively decreases toward the last layers, reaching very low values in comparison to the embedding dimension. In addition, the evolution of the probability density of neighbors as measured for ImageNet~\citep{russakovsky2015imagenet} on several CNN architectures shows almost no overlap with the output and ground-truth distributions throughout most layers. Specifically, an abrupt overlap emerges in a ``nucleation''-type process occurring at layer $142$ of ResNet152 (i.e., toward the final layers of the network). Our exploration adds to this understanding that deep models feature a step-like mean absolute principal curvature profile. For the majority of layers, mean curvature and curvature distribution remain relatively fixed and small in absolute values (Figs.~\ref{Fig:common_mapc}, \ref{Fig:pc_distrib}). In contrast, a sharp increase in curvature appears in the final layers of the network. Combining our findings with previous work, we obtain a more comprehensive picture of the data manifold: during the first layers, the network maintains almost flat manifolds, allowing samples to move freely across layers as more directions are available (flat MAPC and high ID). Then, as computation proceeds, samples concentrate near their same-class samples in highly-curved peaks, facilitating separation between clusters. This understanding can be utilized by designing model whose curvature profile is step-like by construction. To conclude, we hope that our analysis in this work will inspire others to further our understanding on data manifolds learned with deep neural networks, allowing to develop better and more sophisticated learning models in the future.

\section{Acknowledgements}
\label{Sec:acknowledgement}
This research was partially supported by the Lynn and William Frankel Center of the Computer Science Department, Ben-Gurion University of the Negev, an ISF grant 668/21, an ISF equipment grant, and by the Israeli Council for Higher Education (CHE) via the Data Science Research Center, Ben-Gurion University of the Negev, Israel.

% cohen and sompolinsky show an increase in classification capacity in relation to a decrease in manifold dimension and radius

% in this context, our analysis shows that the network behaves in an opposing manner: the data manifold curvature increases with depth whereas the decision boundary curvature flattens with depth; it is like the space needs to be "wrinkled" in order to match a flat boundary

% an illustration of a spiky sphere, relation to nucleation paper; Euclidean space | minimal surface | sphere

% Recent empirical studies on the evolution of probability densities in deep convolutional networks find that the neighborhood of a data point changes suddenly, toward the final layers of a ResNet152 network~\citep{doimo2020hierarchical}. Our results are consistent with these studies, providing a geometrical perspective to the existing distributional understanding. In particular, our curvature analysis provides an explanation for the nucleation 

\clearpage
\bibliography{refs}
\bibliographystyle{icml2023}

%%%%%%%%%%%%%%%%%%%%%%%%%%%%%%%%%%%%%%%%%%%%%%%%%%%%%%%%%%%%%%%%%%%%%%%%%%%%%%%
%%%%%%%%%%%%%%%%%%%%%%%%%%%%%%%%%%%%%%%%%%%%%%%%%%%%%%%%%%%%%%%%%%%%%%%%%%%%%%%
% APPENDIX
%%%%%%%%%%%%%%%%%%%%%%%%%%%%%%%%%%%%%%%%%%%%%%%%%%%%%%%%%%%%%%%%%%%%%%%%%%%%%%%
%%%%%%%%%%%%%%%%%%%%%%%%%%%%%%%%%%%%%%%%%%%%%%%%%%%%%%%%%%%%%%%%%%%%%%%%%%%%%%%
\clearpage
\appendix
\onecolumn
% comparison of MAPC vs. MAMC vs. MARC (mean absolute principal curvatures to mean absolute mean curvatures to mean absolute riemann curvatures)
\section{Comparing Different Metrics of Curvature}
\label{App:cmp_curv_metrics}
\begin{figure}[ht]
        \centering
        \includegraphics[width=1\linewidth]{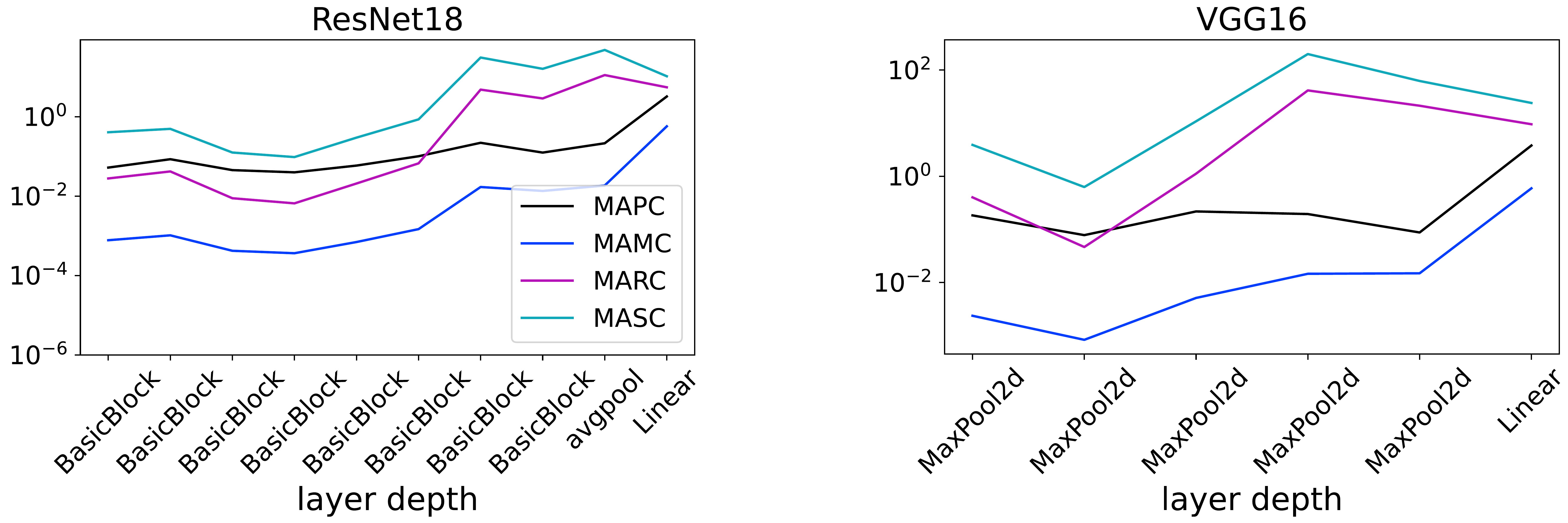}
        \caption{\textbf{Comparison of different curvature metrics: MAPC, MAMC, MARC and MASC.}}
        \label{Fig:cmp_curv_metrics}
    \end{figure}

The results shown in this paper measure curvature by investigating the Mean Absolute Principal Curvature (MAPC), which is given by the average of the absolute values of eigenvalues of the estimated Hessian matrices. To perform a comprehensive analysis, we show the behaviour of three additional metrics that measure curvature. \textbf{Mean Absolute Mean Curvature} (MAMC) computes the mean absolute value on the mean curvature, which is the natural extension of mean curvature of surfaces to manifolds in higher dimensions. The mean curvature is defined as the mean principal value, of the Hessian matrix. We compute the mean curvature for each one of the $\alpha=1,\dots,D-d$ Hessian matrices and then take the mean of their absolute values. \textbf{Mean Absolute Riemann Curvature} (MARC) computes the mean of the absolute value of all the components in the Riemann curvature tensor. \textbf{Mean Absolute Sectional Curvature} computes the mean of the absolute value of the sectional curvatures. As shown in Fig.~\ref{Fig:cmp_curv_metrics}, the pairs MAPC, MAMC and MARC, MASC show a similar trend while MARC and MASC are larger consistently across different networks. Overall, all the metrics exhibit comparable behaviours and due to the lack of a canonical metric for providing a single scalar value that represents the curvature of a manifold, we opted to use MAPC.

% waiting for mean absolute riemann curvature data

% \section{Curvature Profile of Untrained Networks}
% \label{App:untrained_supp}
% To augment the results of Sec.~\ref{Subsec:untrained}, we added Fig.~\ref{Fig:mapc_cmp_random} which shows the MAPC profile for ResNet50 and Resnet101 for untrained networks in comparison to VGG architectures (Fig.~\ref{Fig:mapc_cmp_random}A) and in comparison to trained ResNet models (Fig.~\ref{Fig:mapc_cmp_random}B). It is clear that untrained networks exhibit a different MAPC characteristic, most notably by the sharp decrease in the last layer.

% \begin{figure}[ht]
%     \centering
%     \begin{overpic}[width=1\linewidth]{figures/mapc_cmp_random.pdf}
%        \put(0, 30){A} \put(55, 30){B}
%     \end{overpic}
%      \caption{\textbf{Comparison of MAPC profiles for baseline models with untrained networks.} \textbf{A)} MAPC graphs for VGG and ResNet neural architectures before training. \textbf{B)} MAPC graphs for ResNet50 and ResNet101 networks before and after training.}
%     \label{Fig:mapc_cmp_random}
% \end{figure}
% \begin{figure}[hb]
%     \centering
%     \begin{overpic}[width=1\linewidth]{figures/mapc_rand_n_reg}
%         \put(0, 30){A} \put(55.5, 30){B}
%     \end{overpic}
%     \caption{\textbf{Comparison of MAPC profiles for ResNet50 and ResNet101 models with untrained networks.}}
%     \label{Fig:mapc_rand_n_reg}
% \end{figure}

\section{Generalization on ImageNet}
\label{App:gen_imagenet}
To verify the generality of our results we analyzed the behavior of curvature along the layers of Resnet models trained on Tiny Imagenet~\cite{le2015tiny} and Imagenet~\citep{deng2009imagenet}. The following sections describe the experiments and results.

\subsection{Tiny ImageNet}
\label{App:tiny_imagenet}
\begin{figure}[!t]
    \centering
    \begin{overpic}[width=1\linewidth]{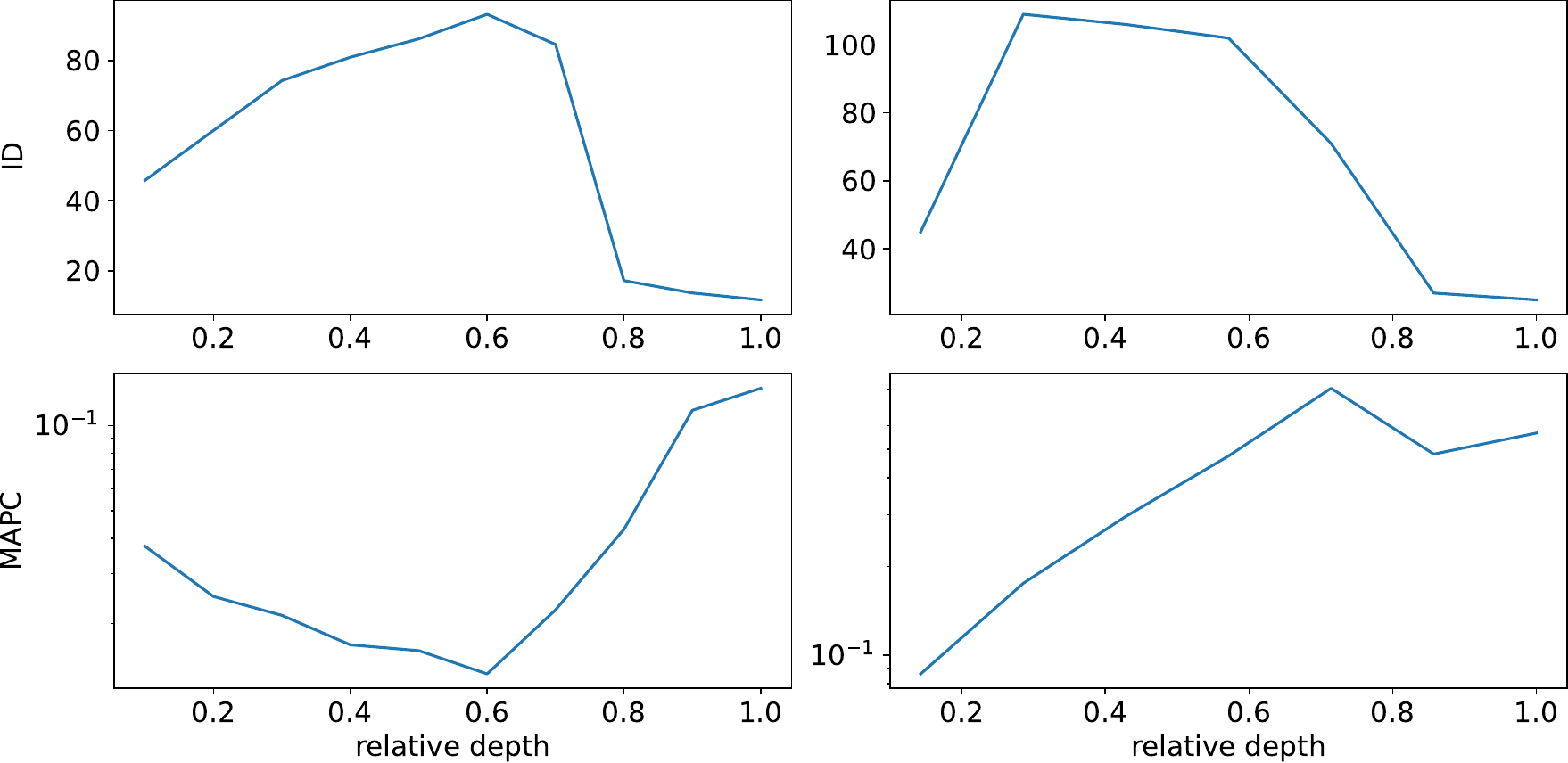}
       \put(0, 48){A} \put(52, 48){B}
    \end{overpic}
    \vspace{-3mm}
    \caption{\textbf{The intrinsic dimension and mean absolute principal curvature.} \textbf{A)} ID and MAPC along the layers of ResNet18 trained on Tiny ImageNet \textbf{B)} ID and MAPC along the layers of ResNet50 trained on ImageNet}
    \label{Fig:tiny_imagenet_mapc}
\end{figure}

\begin{figure}[!t]
    \centering
    \begin{overpic}[width=.5\linewidth]{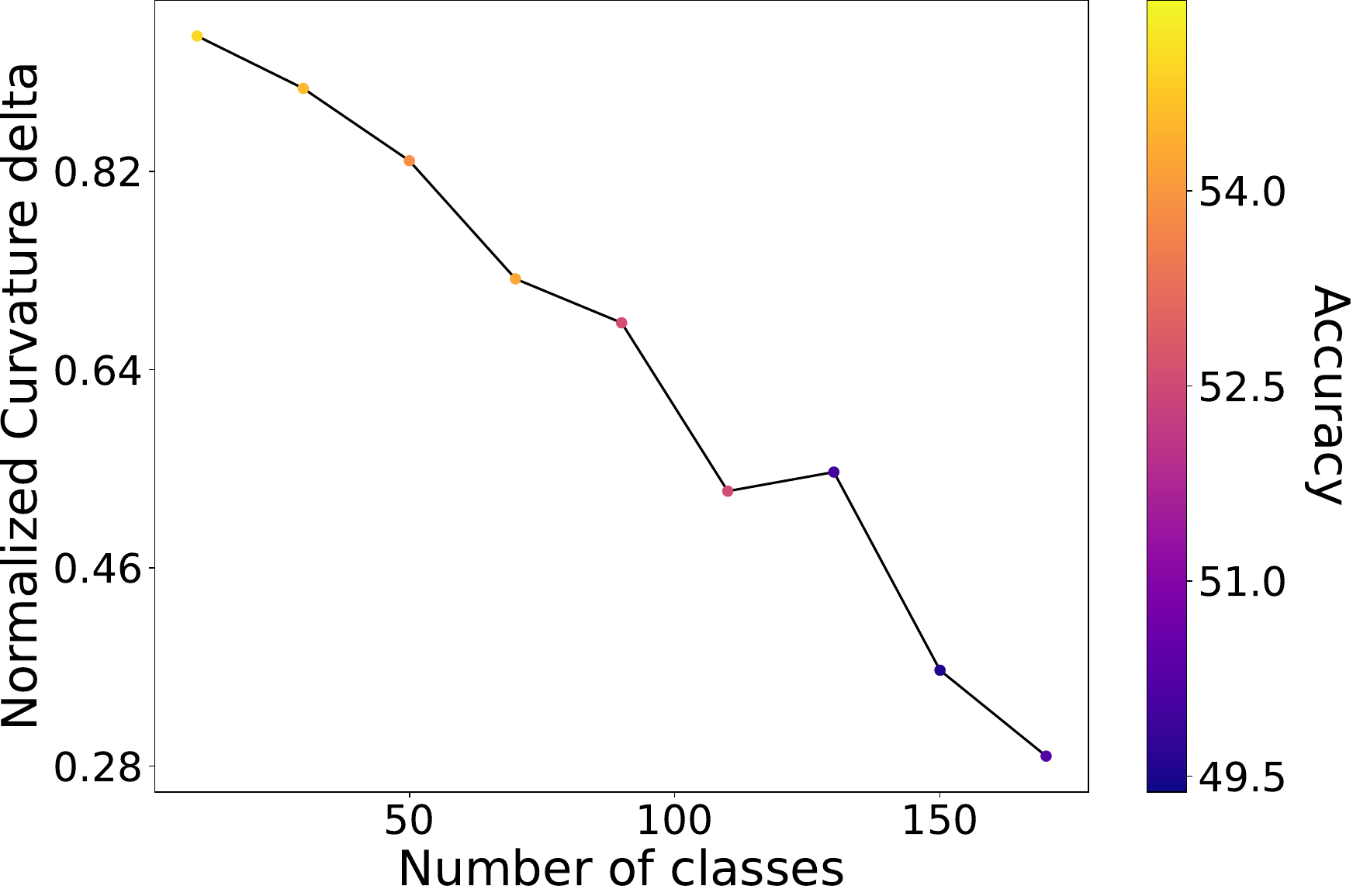}
       % \put(0, 30){A} \put(55, 30){B}
    \end{overpic}
    \vspace{-3mm}
    \caption{\textbf{NMAPC gap with respect to the number of classes used for training.} The color bar represents the achieved accuracy. There is a correlation between the NMAPC gap and accuracy value}
    \label{Fig:tiny_imagenet_delta}
\end{figure}

We trained ResNet18 models on the Tiny Imagenet dataset. The latter dataset is a subset of Imagenet containing $100$k images of $200$ classes ($500$ images per class), downsized to $64 \times 64$ colored images. We computed the intrinsic dimension and MAPC profiles, and we show the results in Fig.~\ref{Fig:tiny_imagenet_mapc}A. Importantly, the MAPC profile is extremely similar to the profiles we demonstrated for CIFAR10 and CIFAR100 on the same architecture. Additionally, we show that the NMAPC gap remains an indicator of the generalization ability. We computed the NMAPC gap with respect to the number of classes used for training Fig.~\ref{Fig:tiny_imagenet_delta}. Overall, we see a strong correspondence between the model performance and the NMAPC gap, similar to the results attained on CIFAR10/CIFAR100.

\subsection{ImageNet}
\label{App:imagenet}

In addition to our experiments on Tiny ImageNet, we also estimated the curvature of $100$ different classes from the ImageNet dataset for a total of $20$k images. The profile we obtained on Imagenet Fig.~\ref{Fig:tiny_imagenet_mapc}B shares several of the key observations we made in the paper. In particular, our Imagenet MAPC profile is generally increasing across layers, it is not correlated with the intrinsic dimension, and it presents a (mild) jump in curvature at the last layer. These results generally align with the claims we made in our paper.

% \begin{figure}[!t]
%     \centering
%     \begin{overpic}[width=1\linewidth]{figures/img_net_res50}
%        % \put(0, 30){A} \put(55, 30){B}
%     \end{overpic}
%     \vspace{-3mm}
%     \caption{\textbf{The intrinsic dimension and mean absolute principal curvature along layers of ResNet50 trained on ImageNet:} \textbf{Left:} Intrinsic dimension. \textbf{Right} Mean absolute principal curvature.}
%     \label{Fig:imagenet_res50}
% \end{figure}

In conclusion, while there are some differences between the MAPC profile on Imagenet in comparison to CIFAR10, CIFAR100, and Tiny Imagenet, the majority of our analysis and observations apply to all these different datasets, extending across multiple architectures, models, and training protocols.

\section{Intrinsic Dimension Estimators Effect on MAPC}
\label{App:id_est_on_mapc}

\begin{figure}[!t]
    \centering
    \begin{overpic}[width=1\linewidth]{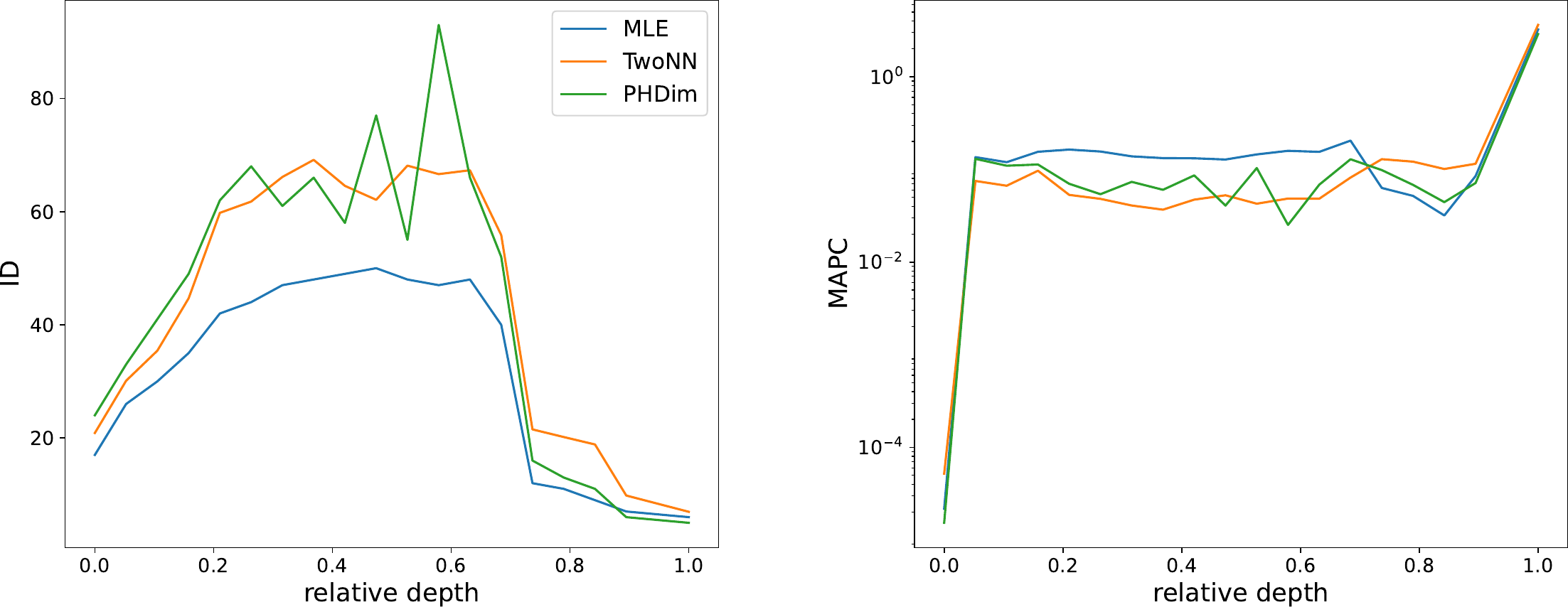}
       \put(0, 37){A} \put(52, 37){B}
    \end{overpic}
    \vspace{-3mm}
    \caption{\textbf{The effect of different ID estimation tools on curvature: A)} the ID values may vary significantly while the MAPC profile remains stable and consistent with our previous results. \textbf{B)} MAPC values do not correlate with the ID values.}
    \label{Fig:id_est_effect_mapc}
\end{figure}

To strengthen our claim that curvature is not necessarily correlated with dimensionality, we estimated the ID of latent data representations computed with ResNet50 on CIFAR10 using the following methods: TwoNN~\citep{facco:SR:2017:twonn}, Maximum Likelihood Estimation (MLE)~\citep{popeintrinsic}, and  Persistent Homology Dimension (PHDim)~\citep{birdal2021intrinsic}. We then used the resulting ID values to estimate the curvature. We show the intrinsic dimension and MAPC shown in Fig.~\ref{Fig:id_est_effect_mapc}. Note that the ID values may vary significantly while the MAPC profile remains stable and consistent with our previous results. Further, note that the MAPC values do not correlate with the ID values. For instance, all ID profiles present a significant drop in values toward the last layer, whereas the MAPC profiles do not change much. In conclusion, these results further strengthen our claim that ID and MAPC are not correlated.

\section{Data Density}
\label{App:data_density}
\begin{figure}[ht]
    \centering
    \includegraphics[width=\linewidth]{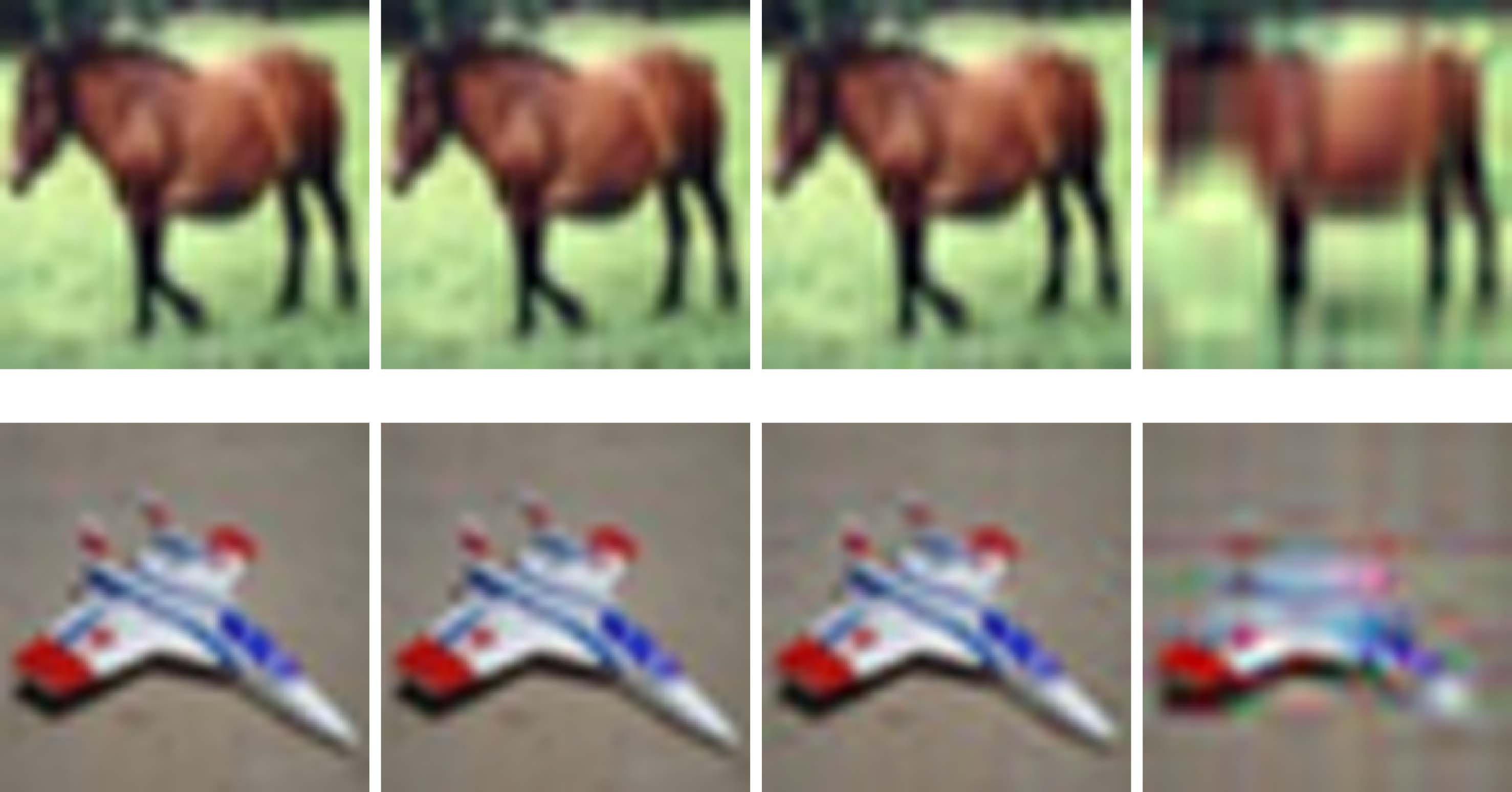}
    \caption{\textbf{Visualization of the neighborhood generation process.} The first column to the left shows a sample from the CIFAR10 data set. Each consecutive column shows the generated image using the SVD method where the number of singular values which were set to zero increases from left to right. Note that nullifying a small amount singular values results in an image that is almost identical to the original image (2nd and 3rd columns), while nullifying more singular values creates noticeable differences (4th column).}
    \label{Fig:svd_close}
\end{figure}

Curvature estimates for high-dimensional and sparse point clouds are extremely noisy and unreliable. To alleviate this issue, we aimed to (locally) increase the density of the data manifold. Our choice
to use SVD is well-motivated from a differential geometry viewpoint. Specifically, the SVD procedure we
described in Sec.~\ref{Sec:Method} is closely-related to computing a first order approximation of the manifold at a point,
and sampling at the neighborhood of the point. Sampled points may slightly deviate from the data
manifold, yet the deviation can be bounded by the absolute value of the modified singular values (which are
close to zero in practice). There are several works~\citep{donoho2003hessian,zhang2004principal,singer2012vector,tyagi2013tangent} that justify the usage of SVD for estimating the tangent plane of a manifold at a given point $p$.
In addition to the theoretical justification we provide for the SVD procedure, we investigated the proximity of the generated neighborhood using affine transformations, k-nearest-neighbors and SVD, see Fig.~\ref{Fig:neigh_dists} in the main text. The affine transformations include rotations in the range of $[-10,10]$ degrees, shear parallel to the x and y axis in the range $[-10,10]$ degrees, horizontal translation in the range $[-.1w,.1h]$ where $w, h$ are the image width and height, and vertical translation in the range $[.1w,.1h]$. Using smaller values for the affine transformation parameters caused the curvature estimation algorithm to fail. It is notable that the generated images using the SVD method create samples that are closer in an Euclidean distance sense along all layers. Visually, the samples generated using the SVD method look almost identical to the original image from which they were generated as can be seen in Fig.~\ref{Fig:svd_close}.

% we also experimented with up-sampling approaches based on standard image transforms, e.g., affine transformations, including translation, shear and rotation (see Fig.~\ref{Fig:mapc_affine}). That is, we train the networks as before, but during inference, we feed a point with its neighborhood based on image transforms such as rotations of the image. Remarkably, we find for CIFAR10 on ResNet and VGG architectures a characteristic profile akin to the curvature profiles we report e.g., in Fig.~\ref{Fig:common_mapc}, with one
% qualitative different feature. The curvature profiles associated with image transforms exhibit a significant drop in the penultimate layer, whereas curvature profiles as reported in Fig.~\ref{Fig:common_mapc} do not present this characteristic. We believe it is related to the low-dimensionality of the data (as governed by the image transforms) allowing to form a low curvature manifold in comparison to our SVD-based sampling which yields closer points, but with potentially more intrinsic dimensions, making it harder to encode using very low MAPC values. 

% \begin{figure}[ht]
%     \centering
%     \begin{overpic}[width=1\linewidth]{figures/mapc_img_transforms.pdf}
%        \put(0, 30){A} \put(55, 30){B}
%     \end{overpic}
%     \caption{\textbf{Effect of generating close samples using affine transformations.} \textbf{A)} MAPC profile estimation using affine transformations for local patch generation. \textbf{B)} Comparison of the mean distance between a sample and its neighborhood along the layers of ResNet18.}
%     \label{Fig:mapc_affine}
% \end{figure}

\section{Estimating the Hessian Matrix}
\label{App:taylor_linsolve}

% building the linear set of equations for estimating the hessians
As discussed in Sec.~\ref{Sec:Method} above, we wish to estimate the Hessian per embedding mapping $f^\alpha$ where $\alpha=1,\dots,D-d$. This is done by building a set of linear equations that solves Eq.~\ref{EQN:maptaylor}:
\begin{equation*}
    f^\alpha(u_{i_j}) = f^\alpha(x_i) + (u_{i_j}-x_i)^T \nabla f^\alpha + \frac{1}{2} (u_{i_j}-x_i)^T H^\alpha (u_{i_j}-x_i) + \mathcal{O}(|u_{i_j}|_2^2) \ ,
\end{equation*}
that is $f^\alpha$ is approximated by solving the system $f^\alpha = \Psi X_{i}$, where $X_{i}$ contains the unknown elements of the gradient $\nabla f^\alpha$ and the hessian $H^\alpha$. We define $f^\alpha=\left[f^\alpha\left(u_{i_1}\right), \cdots, f^\alpha\left(u_{i_K}\right)\right]^T$, where $u_{i_j}$ are points in the neighborhood of $x_i$, projected to the local natural orthogonal frame. The local natural orthonormal coordinate frame is defined as the basis associated with the tangent space and normal space at a point $p$ of the manifold. In practice, the coordinate frame is generated using PCA, where the first $d$ coordinates (associated with the most significant modes, i.e., largest singular values) represent the tangent space, and the rest encode the normal space.
Then, we define $\Psi=\left[\Psi_{i_1}, \cdots, \Psi_{i_K}\right]$, where $\Psi_{i_j}$ is given via
\begin{equation*}
    \Psi_{i_j}=\left[u_{i_j}^1, \cdots, u_{i_j}^d,\left(u_{i_j}^1\right)^2, \cdots,\left(u_{i_j}^d\right)^2,\left(u_{i_j}^1 \times u_{i_j}^2\right), \cdots,\left(u_{i_j}^{d-1} \times u_{i_j}^d\right)\right] \ .
\end{equation*}
We solve $f^\alpha=\Psi X_{i}$ by using the least square estimation resulting in $X_{i}=\Psi^{\dagger}f^\alpha$, such that $X_{i}=\left[{\nabla f^\alpha}^1,\cdots, {\nabla f^\alpha}^d, {H^\alpha}^{1,1}, \cdots, {H^\alpha}^{d,d}, {H^\alpha}^{1,2}, \cdots, {H^\alpha}^{d-1,d}\right]$, that is, we estimate only the upper triangular part of ${H^\alpha}$ since it is a symmetric matrix. We do not use the elements of the gradient $\nabla f^\alpha$ for the CAML algorithm, it is only computed as a part of the hessian $H^\alpha$ estimation. We refer the reader for a more comprehensive and detailed discussion in~\citep{li2018curvature}.

\section{Characteristic mean absolute principal curvature}

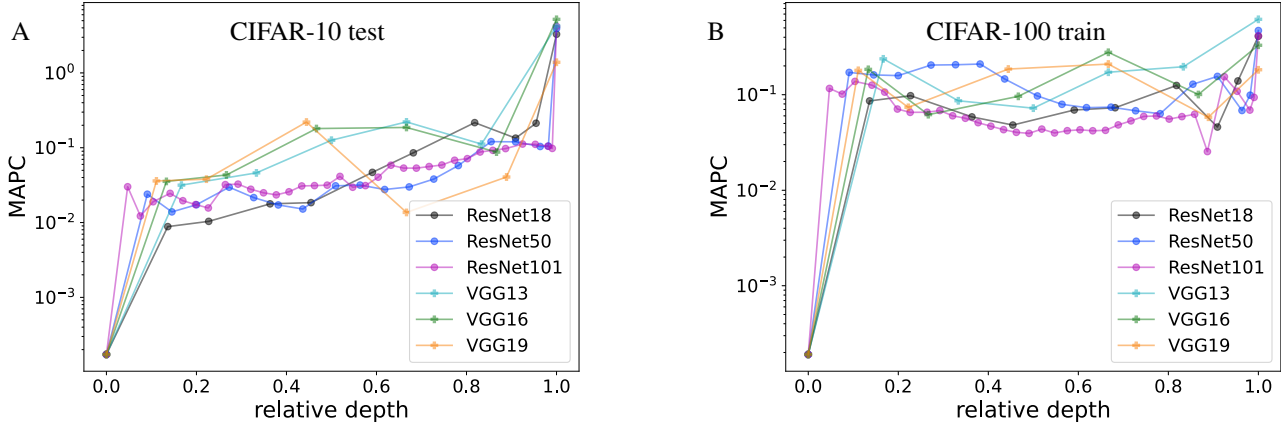
\begin{figure}[ht]
    \centering
    \begin{overpic}[width=1\linewidth]{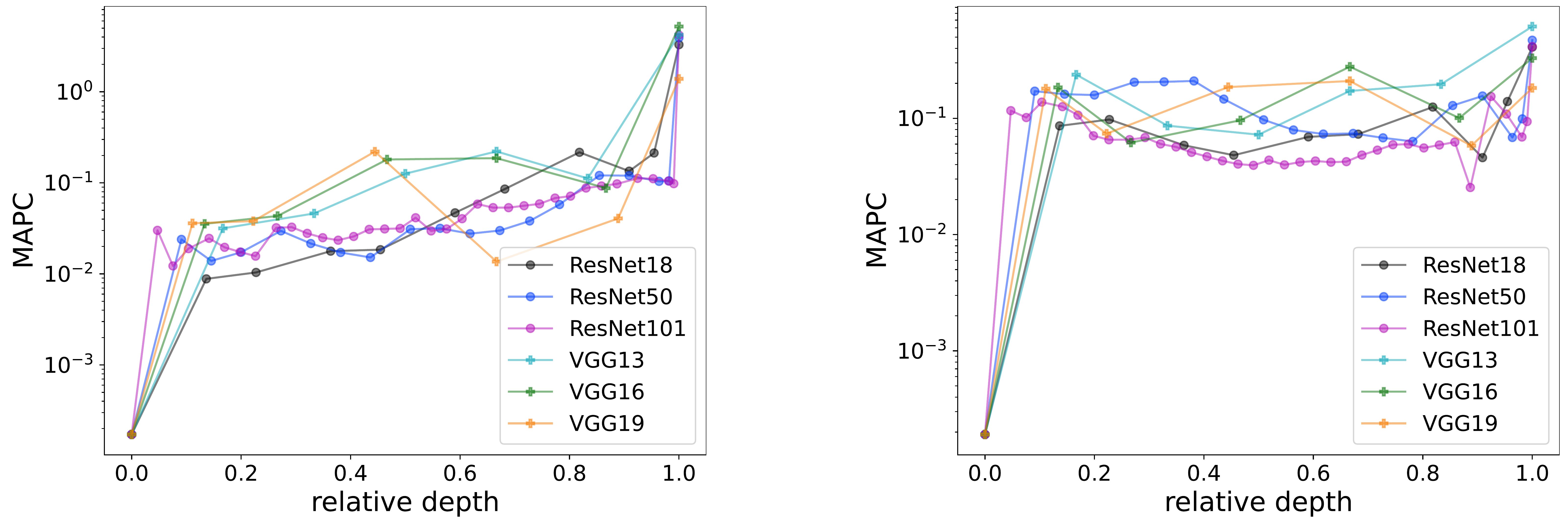}
        \put(1, 30){A} \put(55, 30){B}
       \put(18, 30){CIFAR-10 test} \put(72, 30){CIFAR-100 train}
    \end{overpic}
    \caption{\textbf{MAPC profiles on various models and datasets. A)} MAPC on six different networks on CIFAR-10 test set. \textbf{B)} MAPC on those same models with CIFAR-100 train set.}
    \label{Fig:common_mapc_app}
\end{figure}

We complement the results shown in Sec.~\ref{Subsec:mapc}, and we demonstrate the mean absolute principal curvature profiles for several networks on CIFAR-10 test set and CIFAR-100 train set as shown in Fig.~\ref{Fig:common_mapc_app} in the left and right panels, respectively. In both cases we observe the typical behavior described before: an initial sharp increase, followed by a flat phase, and ending with a final increase. Notably, the maximum MAPC values for CIFAR-100 are \emph{lower} in comparison to both CIFAR-10 train and test sets. Moreover, the gap in the final increase in curvature is \emph{smaller} for CIFAR-100. These results are consistent with our discussion in Sec.~\ref{Subsec:pc_gen_err}.

% \section{Training dynamics} 
% \label{App:train_dynamics}
% In Sec.~\ref{Sec:Results}, we observed a correspondence between model performance and the NMAPC gap. We were interested to see if the training process of the network increases the mentioned gap. We trained a Resnet18 network with CIFAR-10 and observed how the gap changes. We hypothesized that the gap will increase as the network training converges. Remarkably, we indeed find that the NMAPC gap is highly correlated with the behavior of the network during training (Fig.~\ref{Fig:train_dynamics}). Each dot in the plot represents a different snapshot of the model during training, and it is positioned with respect to its accuracy on the test set as a function of the epoch. The points are colored by their NMAPC gap (see colorbar on the right). Overall, we observe that during training the accuracy increase in conjunction with the gap, meaning that the network favors a large gap to increase its performance.

% % TODO: re-generate this plot
% \begin{figure}[ht]
%     \centering
%     \includegraphics[width=.5\linewidth]{./figures/resnet18_train_dynamics.pdf}
%     \caption{\textbf{Training dynamics of the NMAPC gap on ResNet18 and CIFAR-10.} The plot shows how the accuracy changes during training, colored by the normalized curvature gap.}
%     \label{Fig:train_dynamics}
% \end{figure}

\section{Riemannian Geometry Background}
\label{App:geo_background}

This section contains the mathematical background necessary for understanding the curvature estimation process. A vast knowledge in differential geometry is needed to fully comprehend the mathematical background listed below, we will not go in to detail for all the tools we use but rather refer the reader to books on the subject, e.g., \citep{lee2006riemannian, petersen2006riemannian}.

\subsection{Problem Statement}
Let $Y=\{y_1, y_2, \cdots, y_N \}\subset\mathbb{R}^{D}$ be the data on which we want to estimate the curvature. We assume that the data lies on a $d$-dimensional manifold $\mathcal{M}$ embedded in $\mathbb{R}^{D}$ where $d$ is much smaller than $D$, thus, $\mathcal{M}$ can be viewed as a sub-manifold of $\mathbb{R}^{D}$. Will will describe how to compute a second-order local approximation of the embedding map $f: \mathbb{R}^d \rightarrow \mathbb{R}^D$,
\begin{equation}
    y_i = f(x_i) + \epsilon_{i} \ , \quad i=1,\dots,N \ ,
\end{equation}
where $X=\{x_1, x_2, \cdots, x_N\}\subset\mathbb{R}^{d}$ are low-dimensional representations of $Y$, and $\{\epsilon_1, \epsilon_2, \cdots \epsilon_N\}$ are corresponding noises. 
\subsection{Riemannian manifold}
A manifold $\mathcal{M}$ is a topological space that locally resembles Eulidean space near each point. This is particularly useful to our work since the manifold hypothesis states that complex high-dimensional data lies in an intrinsically low-dimensional manifold. \\
%embeding mapping $f: \mathbb{R}^d \rightarrow \mathcal{M}$
\begin{definition} [Tangent Space] {abs}
Let $\mathcal{M} \subset \mathbb{R}^D$ be a manifold where $\mathbb{R}^D$ is the ambient space. For every point $\mathbf{p} \in \mathcal{M}$, a tangent space is a vector space that represents the set of all vectors tangent to given differentiable manifold $\mathcal{M}$ at point $p$, denoted by $T_p \mathcal{M}$.
\end{definition}

\begin{definition} [Riemannian Manifold] A Riemannian manifold $\langle\mathcal{M}, g\rangle$ is a manifold $\mathcal{M}$ endowed with an inner product $g_{\mathbf{p}}$ at the tangent space $T_{\mathrm{p}} \mathcal{M}$ at each point $\mathbf{p}$ that varies smoothly from point to point in the sense that if $X$ and $Y$ are differentiable vector fields on $\mathcal{M}$, then $\mathbf{p} \mapsto g_{\mathbf{p}}(X(\mathbf{p}), Y(\mathbf{p}))$ is a smooth function.
\end{definition}

% Riemann Curvature
\begin{definition} [Riemann Curvature~\citep{petersen2006riemannian}] Let $\langle\mathcal{M}, g\rangle$ be a Riemannian manifold and $\nabla$ the Riemannian connection. The curvature tensor is a $(1,3)-$ tensor defined by
$$
\mathcal{R}(X, Y) Z=\nabla_X \nabla_Y Z-\nabla_Y \nabla_X Z-\nabla_{[X, Y]} Z \ ,
$$
on vector fields $X, Y, Z$.
Using Riemannian metric $g, \mathcal{R}(X, Y) Z$ can be changed to a $(0,4)$-tensor:
$$
\mathcal{R}(X, Y, Z, W)=g(\mathcal{R}(X, Y) Z, W) \ .
$$
\end{definition}

% Local isometry - not sure how it used

% \begin{defn} [Local Isometry]  Let $(\mathcal{M}, g)$ and $(\mathcal{N}, h)$ be two Riemannian manifolds where $g$ and $h$ are Riemannian metrics on them. For a map between manifolds $F: \mathcal{M} \rightarrow \mathcal{N}, F$ is called local isometry if $h\left(d F_p(v), d F_p(v)\right)=$ $g(v, v)$ for all $p \in \mathcal{M}, v \in T_p \mathcal{M}$. Where $d F$ is the differential of $F$.
% \end{defn}
% % Under local isometry, $d F$ is a linear isometry between the corresponding tangent spaces $T_p \mathcal{M}$ and $T_{F(p)} \mathcal{N}$
% Intrinsic geometry such as Riemann curvature tensor and sectional curvature tensor will be preserved under local isometric mapping.

% Sectional Curvature
\begin{definition}  [Sectional Curvature~\citep{petersen2006riemannian}] Let $\langle\mathcal{M}, g\rangle$ be a Riemannian manifold, $p \in \mathcal{M}, u, v \in T_p \mathcal{M}$ are two linearly independent tangent vectors, the sectional curvature of the plane $\mathbb{R} u+\mathbb{R} v$ will be defined as
$$
K(u, v)=\frac{\mathcal{R}(u, v, u, v)}{\langle u, u\rangle\langle v, v\rangle-\langle u, v\rangle^2} \ ,
$$
where $\mathcal{R}$ is the Riemann curvature tensor.
\end{definition}

% estimation of the curvature tensor
\subsection{Computation of the Riemann Curvature Tensor}
Our next task is to compare the Riemannian curvature of $\mathcal{M}$ with that of ambient space $\widetilde{\mathcal{M}}$. According to the definition of curvature tensor, we first give the relationship between the Riemannian connection $\nabla$ of $\mathcal{M}$ and $\widetilde{\nabla}$ of $\widetilde{\mathcal{M}}$:
$$
\tilde{\nabla}_X Y=\nabla_X Y+\mathcal{B}(X, Y),
$$
where the normal component is known as the second fundamental form $\mathcal{B}(X, Y)$ of $\mathcal{M}$. The second fundamental form uncovers the extrinsic structure of a manifold $\mathcal{M}$
relative to ambient space $\widetilde{\mathcal{M}}$. How the manifold is curved with respect to the ambient space is measured by the second fundamental form.
% I dont have a simple/intuitive explanation to what a Riemannian connection is. 

% In Riemannian sub-manifold, one main task is to compare the Riemannian curvature of $\mathcal{M}$ with that of ambient space $\widetilde{\mathcal{M}}$. According to the definition of curvature tensor, we first give the relationship between the Riemannian connection $\nabla$ of $\mathcal{M}$ and $\widetilde{\nabla}$ of $\widetilde{\mathcal{M}}[14]$ :
% $$
% \tilde{\nabla}_X Y=\nabla_X Y+\mathcal{B}(X, Y),
% $$
% where the normal component is known as the second fundamental form $\mathcal{B}(X, Y)$ of $\mathcal{M}$

% Therefore, we can interpret the second fundamental form as a measure of the difference between the Riemannian connection on $\mathcal{M}$ and the ambient Riemannian connection on $\widetilde{\mathcal{M}}$. Based on the relationship between $\nabla$ and $\tilde{\nabla}$, we give the following theorem to show the relationship between the Riemannian curvature of sub-manifold and the Riemannian curvature of ambient space.

% The Gauss Equation
\begin{theorem} [The Gauss Equation~\citep{lee2006riemannian}] For any vector fields $X, Y, Z, W \in$ $T\mathcal{M}$ the tangent bundle of $\mathcal{M}$, the following equation holds:

\begin{align*}
\widetilde{\mathcal{R}}(X, Y, Z, M)=\mathcal{R}(X, Y, Z, W)- 
\langle B(X, W), B(Y, Z)\rangle+\langle B(X, Z), B(Y, W)\rangle
\end{align*}
\end{theorem}
where $\widetilde{\mathcal{R}}$ is the Riemann curvature tensor of $\widetilde{\mathcal{M}}$ and $\mathcal{R}$ is that of $\mathcal{M}$.
Riemannian curvature of the ambient space can be decomposed into two components. In this paper the ambient space is Euclidean space $\mathbb{R}^D$, so $\widetilde{\mathcal{R}}(X, Y, Z, W)=$ 0 . In this case, the Riemannian curvature of $\mathcal{M}$ is represented as:

\begin{align*}
\mathcal{R}(X, Y, Z, W)=\langle\mathcal{B}(X, W), \mathcal{B}(Y, Z)\rangle 
-\langle\mathcal{B}(X, Z), \mathcal{B}(Y, W)\rangle
\end{align*}

To compute the value of the second fundamental form, we construct a local natural orthonormal coordinate frame $\left\{\frac{\partial}{\partial x^1}, \cdots, \frac{\partial}{\partial x^d}, \frac{\partial}{\partial y^1}, \cdots, \frac{\partial}{\partial y^{D-d}}\right\}$ of the ambient space $\widetilde{\mathcal{M}}$ at point $p$, the restrictions of $\left\{\frac{\partial}{\partial x^1}, \cdots, \frac{\partial}{\partial x^d}\right\}$ to $\mathcal{M}$ form a local orthonormal frame of $T_p \mathcal{M})$. The last $D-d$ orthonormal coordinates $\left\{\frac{\partial}{\partial y^1}, \cdots, \frac{\partial}{\partial y^{D-d}}\right\}$ form a local orthonormal frame of $\mathcal{N}_p(\mathcal{M})$. Under the locally natural orthonormal coordinate frame, the embedding map $f$ is redefined as $f\left(x^1, x^2, \cdots, x^d\right)=\left[x^1, x^2, \cdots, x^d, f^1, \cdots, f^{D-d}\right]$, where $x \doteq\left[x^1, x^2, \cdots, x^d\right]$ are natural parameters. Then the second fundamental form $\mathcal{B}$ can be written as: 
\begin{align*}
\mathcal{B}\left(\frac{\partial}{\partial x^i}, \frac{\partial}{\partial x^j}\right)=\sum_{\alpha=1}^{D-d} h_{i j}^\alpha \frac{\partial}{\partial y^\alpha}
\end{align*}
with $h_{i j}^\alpha,(\alpha=1, \cdots, D-d)$ being the second derivative $\frac{\partial^2}{\partial x^i \partial x^i}$ of embedding component function $f^\alpha$, which constitutes the Hessian matrix $H^\alpha=\left(\frac{\partial^2}{\partial x^2 \partial x^j}\right)$, correspondingly, the Riemann curvature tensor of $\mathcal{M}$ is represented as:
\begin{align*}
R_{i l j k}=\sum_{\alpha=1}^{D-d}\left(h_{i k}^\alpha h_{l j}^\alpha-h_{i j}^\alpha h_{l k}^\alpha\right) .
\end{align*}
It follows
that to compute the Riemann curvature of Riemannian submanifold $\mathcal{M}$, we only need to estimate the Hessian matrix of the embedding map $f$. The Hessian
matrix estimation is described in Sec.~\ref{App:taylor_linsolve}.

%%%%%%%%%%%%%%%%%%%%%%%%%%%%%%%%%%%%%%%%%%%%%%%%%%%%%%%%%%%%%%%%%%%%%%%%%%%%%%%
%%%%%%%%%%%%%%%%%%%%%%%%%%%%%%%%%%%%%%%%%%%%%%%%%%%%%%%%%%%%%%%%%%%%%%%%%%%%%%%

\end{document}